\documentclass[11pt]{article}

\usepackage[preprint]{acl}

\usepackage{times}
\usepackage{latexsym}

\usepackage[T1]{fontenc}
\usepackage[utf8]{inputenc}

\usepackage{microtype}

\usepackage{inconsolata}

\usepackage{graphicx}

\usepackage{algorithm}
\usepackage[noend]{algpseudocode}
\usepackage{booktabs}
\usepackage{algpseudocode}
\usepackage[most]{tcolorbox}
\usepackage{placeins}
\usepackage{listings}
\usepackage{amsmath}

\usepackage{xcolor}   % 색상
\newcommand{\cmark}{\ding{51}}%
\newcommand{\xmark}{\ding{55}}%
\usepackage{pifont}   
\usepackage{csquotes}
\usepackage{textgreek}

\usepackage{amsmath}
\usepackage{amsthm}
\usepackage{enumitem}
\usepackage{float}  % preamble
\usepackage[normalem]{ulem} % 취소선
\usepackage{stfloats}
\usepackage{hyperref}       % hyperlinks
\usepackage{url}            % simple URL typesetting
\usepackage{amsfonts}       % blackboard math symbols
\usepackage{nicefrac}       % compact symbols for 1/2, etc.
\usepackage{microtype}      % microtypography
\usepackage{siunitx}
\usepackage{multirow} 
\tcbuselibrary{breakable} 
\usepackage{subcaption}

\usepackage{xurl}
\tcbuselibrary{listings,breakable,skins}

\newtcblisting{mywidebox}[1][]{%
  colback=white,
  colframe=black!75,
  fonttitle=\bfseries,
  title={#1},
  text only,          % ✅ 이것 추가 (아래쪽만 출력)

}

\title{MATA: Multi-Agent Framework for Reliable and Flexible \\ Table Question Answering}

\author{
 \textbf{Sieun Hyeon\textsuperscript{1}}, 
 \textbf{Jusang Oh\textsuperscript{2}}, 
 \textbf{Sunghwan Steve Cho\textsuperscript{1}}, 
 \textbf{Jaeyoung Do\textsuperscript{1,2}}\thanks{Corresponding author}, 
\\
 \textsuperscript{1}Department of Electrical and Computer Engineering, Seoul National University, \\
 \textsuperscript{2}Interdisciplinary Program in Artificial Intelligence, Seoul National University
\\
\textit{\{zxc2692, dhwntkd412, steve97, jaeyoung.do\}@snu.ac.kr}
}

\begin{document}
\maketitle
\begin{abstract}
Recent advances in Large Language Models (LLMs) have significantly improved table understanding tasks such as Table Question Answering (TableQA), yet challenges remain in ensuring reliability, scalability, and efficiency, especially in resource-constrained or privacy-sensitive environments. In this paper, we introduce MATA, a multi-agent TableQA framework that leverages multiple complementary reasoning paths and a set of tools built with small language models. MATA generates candidate answers through diverse reasoning styles for a given table and question, then refines or selects the optimal answer with the help of these tools. Furthermore, it incorporates an algorithm designed to minimize expensive LLM agent calls, enhancing overall efficiency. MATA maintains strong performance with small, open-source models and adapts easily across various LLM types. Extensive experiments on two benchmarks of varying difficulty with ten different LLMs demonstrate that MATA achieves state-of-the-art accuracy and highly efficient reasoning while avoiding excessive LLM inference. Our results highlight that careful orchestration of multiple reasoning pathways yields scalable and reliable TableQA. The code is available at https://github.com/AIDASLab/MATA.
\end{abstract}

\begin{table*}[h]
\centering
\small
\begin{tabular}{p{4.3cm}|p{1.1cm}|p{1.8cm}|p{1.7cm}|p{2.0cm}|p{2.3cm}} 
\toprule
\textbf{Prior Works} &
\textbf{Model-Agnostic} &
\textbf{LLM Call Optimization} &
\textbf{Multi-Stage Verification} &
\textbf{Multiple Metric Analysis} &
\textbf{Multiple Reasoning Method}
\\
\midrule
SynTQA \cite{zhang-etal-2024-syntqa}  & 
\textcolor{red}{\xmark} &  
\textcolor{red}{\xmark} &
\textcolor{red}{\xmark} &
\textcolor{red}{\xmark} &
CoT, SQL  \\

Mix-SC \cite{Mix-SC} &
\textcolor{red}{\xmark} &  
\textcolor{red}{\xmark} &
\textcolor{red}{\xmark} &
\textcolor{red}{\xmark} &
CoT, Python  \\

TabLaP \cite{TabLaP} & 
\textcolor{red}{\xmark} &  
\textcolor{red}{\xmark} &
\textcolor{green}{\cmark} &
\textcolor{red}{\xmark} &
CoT, Python   \\

Chameleon \cite{chameleon} & 
\textcolor{red}{\xmark} &  
\textcolor{red}{\xmark} &
\textcolor{red}{\xmark} &
\textcolor{red}{\xmark} &
CoT,  Python \\

ReAcTable \cite{ReAcTable} & 
\textcolor{red}{\xmark} &  
\textcolor{red}{\xmark} &
\textcolor{green}{\cmark} &
\textcolor{green}{\cmark} &
Python, SQL  \\

Chain-of-Table \cite{wang2024chainoftable} & 
\textcolor{red}{\xmark} &  
\textcolor{red}{\xmark} &
\textcolor{red}{\xmark} &
\textcolor{green}{\cmark} &
CoT, Python  \\

\textbf{MATA} &
\textcolor{green}{\cmark} &  
\textcolor{green}{\cmark} &
\textcolor{green}{\cmark} &
\textcolor{green}{\cmark} &
CoT, Python, SQL  \\

\bottomrule
\end{tabular}
\caption{Comparison with prior works. MATA uniquely addresses five key aspects of TableQA, unlike others.}
\label{tab:opensource}
\end{table*}

\section{Introduction}

Tables serve as a foundational medium for representing structured data, central to data storage, organization, and analytics. Although advances in programming languages and database systems have improved the accessibility of tabular data~\cite{mckinney-proc-scipy-2010, 10.1145/800296.811515}, interacting with and interpreting tables remains a significant challenge for non-technical users who lack coding expertise. This challenge has motivated active research into table understanding tasks, particularly Table Question Answering (TableQA)~\cite{pasupat-liang-2015-compositional, lu2023dynamic, tablebench}, which aims to enable natural language interaction with tables.

Recent progress in Large Language Models (LLMs) has dramatically expanded the possibilities for natural language interfaces to complex tabular information~\cite{zhang2023nli, TableLLM2024,fang2024llms}, promising more accessible and user-friendly solutions for a broader audience. However, practical deployment of TableQA systems presents several persistent challenges that have not been fully addressed in prior work.

First, a critical limitation of prior TableQA methods is the lack of model-agnosticism.
Most high-performing language models are closed-source and accessible only through paid APIs, which poses significant challenges in settings where privacy concerns, data ownership, or cost restrictions make the use of open-source models necessary. While previous studies \cite{wang2024chainoftable, ReAcTable, chameleon, TabLaP, zhang-etal-2024-syntqa, Mix-SC} have demonstrated strong performance with proprietary LLMs~\cite{openai_hello_gpt4o2024, anthropic_claude37_blog2025}, it remains unclear whether comparable results can be achieved using open-source models, particularly those with smaller parameter sizes. In other words, the reliability of TableQA systems in such constrained and practical environments has yet to be thoroughly investigated.

Second, existing frameworks tend to rely on repeated LLM inferences to boost answer reliability, often incurring substantial computational costs. Techniques like Self-Consistency~\cite{wang2023selfconsistency} or Best-of-N~\cite{kang2025scalablebestofnselectionlarge} have improved performance on complex tasks, but excessive LLM calls can lead to diminishing returns, higher expense, and, in some cases, even degrade accuracy due to over-prompting~\cite{huang2025is, NEURIPS2024_51173cf3}. An optimal TableQA framework must judiciously manage the number of LLM inferences, balancing accuracy with efficiency, which is a requirement that previous work has rarely addressed in depth.

Third, while LLMs support a range of reasoning strategies, including Chain-of-Thought (CoT)~\cite{CoT}, Program-of-Thought (PoT)~ \cite{PoT}, and text-to-SQL (text2SQL)~\cite{zhong2017seq2sqlgeneratingstructuredqueries}, most existing TableQA frameworks fail to exploit this full diversity. Previous studies~\cite{Mix-SC, TabLaP, zhang-etal-2024-syntqa} have shown that text-based reasoning (e.g., CoT) may excel for ambiguous or intuitive queries, while code-based (e.g., PoT) or SQL-based (e.g., Text2SQL) approaches can provide greater precision in numerical reasoning. However, the relative advantage of each reasoning path depends heavily on the characteristics and training data of the underlying model (see Appendix~\ref{appendix:A}), and no single approach is universally optimal. A robust TableQA system should therefore generate and validate answers through multiple complementary reasoning paths to maximize reliability across different models and scenarios.

Incorporating all these considerations, in this paper, we propose a novel TableQA framework, MATA (\underline{M}ulti-\underline{A}gent Framework for \underline{T}ableQ\underline{A}). MATA orchestrates diverse reasoning strategies using a coordinated set of LLM agents and lightweight tools. Its architecture includes mechanisms for multi-step answer verification and an efficient scheduling algorithm that minimizes unnecessary LLM calls, thereby enhancing both reliability and efficiency.

%%%%%%%%% CR Add  %%%%%%%%%
The key distinction of MATA is that reasoning diversity does not require a fixed inference budget. Instead of always executing all reasoning paths, MATA uses lightweight controllers to decide which CoT, PoT, and text2SQL branches are necessary and when verification can stop.
%%%%%%%%% CR Add  %%%%%%%%%

To rigorously evaluate the model-agnostic capabilities and practical effectiveness of MATA, we conduct comprehensive experiments with ten different language models, spanning both small with up to 10B parameters, and large models with more than 10B parameters. In contrast to prior work, we employ three complementary evaluation metrics (i.e., exact match, fuzzy matching, and token-level F1) to more accurately reflect the varied output formats of LLMs in TableQA tasks. 

Notably, MATA achieves up to 40.1\% improvement in exact match, 21.9\% in fuzzy matching, and 33.1\% in F1 score over the strongest baseline on a challenging benchmark, underscoring its superior effectiveness. Importantly, these gains are not simply a result of indiscriminately increasing or minimizing the number of LLM inference calls. Instead, MATA attains high accuracy by dynamically invoking an appropriate number of inference steps—striking a careful balance between reasoning diversity and computational efficiency, rather than blindly maximizing or minimizing LLM usage compared to other baselines. Our main contributions are as follows:

% \noindent \textbf{(1)} We propose MATA, a model-agnostic TableQA framework that unifies Chain-of-Thought, Program-of-Thought, and text-to-SQL reasoning within a coordinated multi-agent system, enabling the generation and rigorous verification of diverse candidate answers for each question.

% \noindent \textbf{(2)} MATA demonstrates consistently strong performance across a broad range of language models, including both closed- and open-source LLMs of varying sizes, thereby overcoming the limitations of prior approaches that depended primarily on large proprietary models.

% \noindent \textbf{(3)} To achieve both high accuracy and computational efficiency, MATA incorporates an intelligent LLM call optimization mechanism and expands evaluation beyond the exact-match accuracy by introducing additional metrics that capture flexible and semantically relevant answer formats.

\noindent \textbf{(1)} We propose MATA, a model-agnostic multi-agent framework that integrates Chain-of-Thought, Program-of-Thought, and text-to-SQL for robust answer generation and verification.

\noindent \textbf{(2)} MATA demonstrates strong performance across three evaluation metrics on diverse open-source and proprietary LLMs, overcoming the reliance on large-scale models. Furthermore, it achieves both accuracy and efficiency through intelligent inference optimization.

\begin{figure*}[t]
\centering
\includegraphics[width=\linewidth]{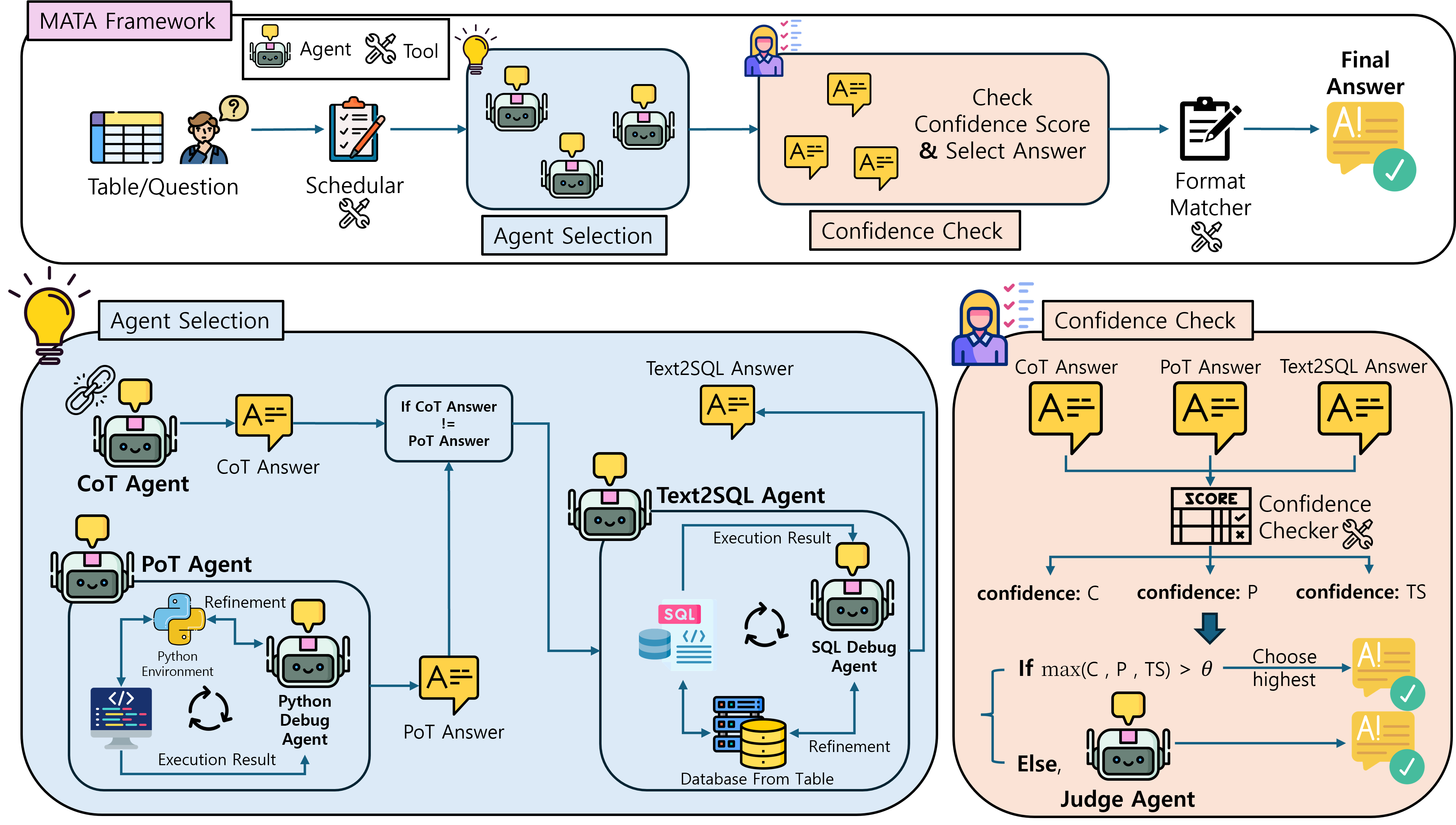} 
\caption{Overview of MATA. The current situation is when the Scheduler(\textit{Sch}) selected PoT first. MATA integrates three complementary reasoning methods (CoT, PoT, text2SQL) through a multi-agent workflow. The Scheduler (\textit{Sch}) prioritizes PoT or text2SQL reasoning based on the table and question, with CoT executed simultaneously. Candidate answers are evaluated by the Confidence Checker (\textit{CC}); if no candidate meets the confidence threshold, the Judge Agent (\textit{JA}) verifies the final answer. The Format Matcher (\textit{FM}) ensures answers are concise.}
\label{fig:MATA}
\end{figure*}

\section{Related Work}

Earlier studies utilized Transformer-based architectures \cite{NIPS2017_3f5ee243} and Pre-trained Language Models (PLMs).  Concurrently, PLMs such as DeBERTaV3 \cite{he2023debertav3}, and others \cite{lewis-etal-2020-bart, 2020t5, brown2020languagemodelsfewshotlearners} enhanced performance by fine-tuning on table-specific or SQL datasets \cite{patnaik2024cabinet, zhang-etal-2024-tablellama, tableGPT-sigmod, liu2022tapex, jiang-etal-2022-omnitab, zhao-etal-2022-reastap, gu-etal-2022-pasta}. Despite promising results, these methods typically underperform LLM-based approaches on out-of-distribution data, though their value in demonstrating PLM capabilities for tabular understanding remains significant.

To mitigate LLM limitations in numerical reasoning, recent methods adopted parallel reasoning paths \cite{Mix-SC, zhang-etal-2024-syntqa, TabLaP}, integrating text-based (e.g., CoT) and code-based reasoning (e.g., PoT, text2SQL). Final answers are selected from candidates generated via textual inference or executed code, often aided by self-consistency mechanisms, external LLM evaluators, or verification models \cite{Mix-SC, zhang-etal-2024-syntqa, LEVER}. Other strategies explored table decomposition into sub-questions \cite{DATER}, iterative reasoning \cite{wang2024chainoftable, ReAcTable}, and external tool integration \cite{chameleon}. Additionally, some methods involve generating and refining SQL or Python code using the same LLM to produce final answers \cite{Binder, self-refine}. Nevertheless, many of these approaches heavily rely on proprietary LLMs, with limited evaluation using open-source models.

\section{Methodology}
\subsection{Preliminary}
\label{sec:preliminary}

In this section, we delineate the components and modules comprising our framework, MATA, while defining the notations employed throughout the paper. The modules within MATA are categorized into two groups: \textbf{Tools} and \textbf{Agents}.

\textbf{Tools} refer to small language models with parameter sizes under 500M. Given that all LLMs used in our experiments possess at least 3B parameters, these tools are markedly smaller in comparison, thereby incurring minimal inference overhead. MATA leverages three such tools.

The Scheduler (\textit{Sch}) serves as the inaugural module, processing the input Table (\textit{T}) and Question (\textit{Q}). Based on several features extracted from \textit{T} (e.g., table size, schema, data types) as well as the semantic meaning of the input \textit{Q}, \textit{Sch} determines whether to perform PoT or text2SQL reasoning first. It is implemented using a lightweight model combining MobileBERT \cite{mobilebert} and a two-layer MLP, totaling 24.65M parameters.

The Confidence Checker (\textit{CC}) module ingests \textit{T}, \textit{Q}, and the candidate answers derived from CoT, PoT, and text2SQL reasoning. It assigns a confidence score to each candidate. Prior research \cite{gu-etal-2022-pasta} has shown that DeBERTaV3 \cite{he2023debertav3} achieves strong performance in table understanding. Accordingly, \textit{CC} is built by fine-tuning DeBERTaV3-large, which has approximately 435M parameters. For more implementation details of the Scheduler and the Confidence Checker, refer to Appendix~\ref{appendix:B}.

The Format Matcher (\textit{FM}) is the final tool in the pipeline. It converts long, verbose responses into concise entities. \textit{FM} is implemented using the qwen2.5-instruct model with 0.5B (500M) parameters, used without any fine-tuning.

\textbf{Agents} denote LLM-based modules with at least 3B parameters. In our experiments, all agents share the same backbone LLM and differ only in their role-specific prompts; \textit{Sch}, \textit{CC}, and \textit{FM} are separate lightweight tool models. MATA incorporates a total of six agents. Specifically, the CoT Agent (\textit{CoTA}), PoT Agent (\textit{PoTA}), and text2SQL Agent (\textit{t2SA}) are responsible for performing CoT, PoT, and text2SQL reasoning, respectively. Owing to the inherent propensity for syntactic errors in code generation, \textit{PoTA} and \textit{t2SA} are supplemented by the Python Debug Agent (\textit{PDA}) and SQL Debug Agent (\textit{SDA}), which rectify outputs from their respective counterparts. Finally, the Judge Agent (\textit{JA}) aggregates candidate answers from CoT, PoT, and text2SQL to adjudicate the ultimate response.

For brevity, subsequent sections will reference each module by its designated abbreviation.

\subsection{Dataset for \textit{Sch} and \textit{CC}}

We curate a large-scale training dataset \footnote{The dataset is publicly available at \url{https://github.com/AIDASLab/MATA/tree/main/experiments}, complete with documentation and licensing for free research use, to foster advancements in efficient and adaptable TableQA systems.} specifically designed to support the fine-tuning of lightweight tools like the \textit{Sch} and \textit{CC} in multi-reasoning TableQA frameworks. To construct it, we leverage three established public TableQA datasets—WikiTQ \cite{pasupat-liang-2015-compositional}, TabMWP \cite{lu2023dynamic}, and TabFact \cite{Chen2020TabFact}—and conduct inferences using three LLMs: phi4-14B \cite{phi4technicalreport}, Qwen2.5-Coder-14B \cite{qwen25codertechnicalreport}, and CodeLLaMA-13B \cite{codellama}. These inferences span three complementary reasoning modes: CoT, PoT, and text2SQL, yielding a total of 173,664 samples from 57,888 unique table-question (\textit{T}, \textit{Q}) pairs, with each pair generating outputs and correctness labels from all three reasoning modes.

\textit{Sch} was trained by labeling each (\textit{T}, \textit{Q}) pair based on whether the PoT or text2SQL path produced the correct answer. \textit{CC} was trained using labels indicating whether each of the CoT, PoT, and text2SQL paths yielded a correct answer.

\subsection{MATA}

MATA obtains the optimal answer \textit{A} for the given table \textit{T} and question \textit{Q} through a 3-stage process. The overall workflow is illustrated in Figure~\ref{fig:MATA}, and the complete algorithm is provided in the Appendix. (See Algorithm~\ref{algo:mata}).

\begin{algorithm}[t]
\small
\caption{Agent Selection with Scheduler(\textit{Sch})}
\begin{algorithmic}

\State $sol_{cot}$, $A_{cot}$ $\gets$ $\textit{CoTA}(T, Q)$

\If{Use\_Scheduler == True}
    \State prob\_pot, prob\_sql $\gets$ $\textit{Sch}(T, Q)$
    \If{prob\_pot \textgreater= prob\_sql}
        \State $\mathcal{C}_{pot}, A_{pot} \gets$ $\text{Code\&Debug}$(\textit{PoTA}, $T$, $Q$, $N$)
        \If{$A_{cot} \neq$ Last Answer in $\mathcal{A}_{pot}$}
            \State $\mathcal{C}_{sql}, A_{sql} \gets$ $\text{Code\&Debug}$(\textit{t2SA}, $T$, $Q$, $N$)
        \Else
            \State $\mathcal{C}_{sql}, A_{sql} \gets$ $\emptyset$
        \EndIf
    \Else
        \State $\mathcal{C}_{sql},  A_{sql} \gets$ $\text{Code\&Debug}$(\textit{t2SA}, $T$, $Q$, $N$)
        \If{$A_{cot} \neq$ Last Answer in $A_{sql}$}
            \State $\mathcal{C}_{pot}, A_{pot} \gets$ $\text{Code\&Debug}$(\textit{PoTA}, $T$, $Q$, $N$)
        \Else
            \State $\mathcal{C}_{pot}, A_{pot} \gets$ $\emptyset$
        \EndIf
    \EndIf
\Else
    \State $\mathcal{C}_{pot}, A_{pot} \gets$ $\text{Code\&Debug}$(\textit{PoTA}, $T$, $Q$, $N$)
    \State $\mathcal{C}_{sql},  A_{sql} \gets$ $\text{Code\&Debug}$(\textit{t2SA}, $T$, $Q$, $N$)
\EndIf

\State \Return $sol_{cot}$, $A_{cot}$, $\mathcal{C}_{pot}$, $A_{pot}$, $\mathcal{C}_{sql}$, $A_{sql}$
\end{algorithmic}
\label{alg:ansselect}
\end{algorithm}

\paragraph{\textbf{Agent Selection with Scheduler(\textit{Sch})}} As illustrated in Algorithm~\ref{alg:ansselect}, upon receiving an input table (\textit{T}) and question (\textit{Q}), MATA employs the Scheduler (\textit{Sch}) to determine the execution priority between the PoT agent (\textit{PoTA}) and the text2SQL Agent (\textit{t2SA}). Concurrently, the CoT Agent (\textit{CoTA}) conducts text-based reasoning to produce both the answer and the associated reasoning path. We denote the CoTA solution text as $sol_{cot}$ and its extracted answer as $A_{cot}$; both are passed to \textit{CC} and, when needed, \textit{JA}. In contrast to \textit{PoTA} and \textit{t2SA}, \textit{CoTA} forgoes additional refinement, as code-based reasoning benefits substantially from debugging, whereas text-based reasoning exhibits marginal improvements even with supplementary LLM inferences (refer to Appendix~\ref{appendix:E}). Consequently, \textit{CoTA} is invoked only once to preserve computational efficiency.

Subsequently, the agent selected by \textit{Sch}—either \textit{PoTA} or \textit{t2SA}—is executed. If its output aligns with the answer generated by \textit{CoTA}, the remaining agent is omitted, and the final answer selection proceeds using only these two candidates.

If the user chooses not to use \textit{Sch}, MATA executes both \textit{PoTA} and \textit{t2SA} without prioritization, thereby generating candidate answers from all three reasoning pathways. This approach, however, elevates the LLM inference overhead, presenting a inherent trade-off between comprehensiveness and efficiency.

\begin{algorithm}[t]
\small
\caption{Code Generation\&Debugging} % (Code\&Debug)
\begin{algorithmic}
% \Require Table ($T$), Question ($Q$), Refinement Count ($N$),
% \midrule
\Function{\textnormal{Code\&Debug}}{$\textit{Agent}, T, Q, N$}
    \State Initialize $\mathcal{C}, A \gets \emptyset$
    \If {$\textit{Agent}$ == \textit{PoTA}}
        \State $\textit{Debug}$ $\gets$ \textit{PDA}
    \Else \State $\textit{Debug}$ $\gets$ \textit{SDA}
    \EndIf

    \State $code^0$, $A^0$ $\gets$ $\textit{Agent}(T, Q)$
    \State Append $(code^0, A^0)$ to $\mathcal{C}, A$
    \For{$i = 0$ to $N-1$}
        \State $code^{i+1}$, $A^{i+1}$ $\gets$ $\textit{Debug}(T, Q, code^i, A^i)$
        \State Append $(code^{i+1}, A^{i+1})$ to $\mathcal{C}, A$
        \If{$\mathrm{Stop\_condition}$ == True}
            \State \textbf{break}
        \EndIf
    \EndFor
    \State \Return $\mathcal{C}$, $A$
\EndFunction

\end{algorithmic}
\label{alg:codegen}
\end{algorithm}

\paragraph{\textbf{Code Generation\&Debugging}} As previously noted, the \textit{CoTA} is invoked only once to ensure cost efficiency, whereas \textit{PoTA} and \textit{t2SA} undergo an additional debugging phase facilitated by their respective Debug Agents. Algorithm~\ref{alg:codegen} delineates the code generation and debugging workflow.

The \textit{PoTA} and \textit{t2SA}, in conjunction with their corresponding Debug Agents—the Python Debug Agent (\textit{PDA}) and SQL Debug Agent (\textit{SDA})—engage in a unified iterative process. Initially, \textit{PoTA} and \textit{t2SA} generate the requisite code, which is executed to yield either an output or error messages. The generated code, along with its execution results, is then forwarded to the pertinent Debug Agent, which refines and debugs the code. The revised code is re-executed to produce updated outputs or error messages. This iterative debugging loop continues for up to $N$ cycles.

To curtail LLM inference expenses, an early termination criterion is implemented: the loop halts if the newly generated code exhibits substantial similarity to its predecessor and yields identical execution results. All iterations of the code and their associated outputs from this loop are subsequently conveyed to the final answer selection phase as candidate responses. The maximum number of debugging iterations $N$ serves as a hyperparameter. Empirical evaluations indicate that $N = 3$ suffices for optimal performance (refer to Appendix~\ref{appendix:E} for hyperparameter optimization details).

\paragraph{\textbf{Final Answer Decision}}

Algorithm~\ref{alg:finalans} describes this process. The answer candidates generated through interactions among the agents are then passed to the final answer selection phase. In this step, the Confidence Checker (\textit{CC}) computes confidence scores for each reasoning path and its corresponding answer. If the score of at least one candidate exceeds a predefined threshold $\theta$, the system skips the additional inference step by the Judge Agent (\textit{JA}) and selects the highest-scoring answer, thereby avoiding an unnecessary LLM agent call. We set $\theta$ to 0.1 through hyperparameter tuning (see Appendix~\ref{appendix:E}).

If none of the candidates exceed $\theta$, the \textit{JA} is invoked to determine the final answer. The \textit{JA} may consider the confidence scores provided by \textit{CC} or make its own judgment independently. Ultimately, the final answer is chosen through a two-step validation process involving both the \textit{CC} and \textit{JA}.

Additionally, we observed that in some cases, LLMs generate excessively verbose responses as final answers, which is undesirable in TableQA tasks where ground truths are typically short phrases or entities. This issue stems from the backbone LLMs’ limited instruction-following ability and can degrade performance. To address this, we introduce a lightweight Format Matcher (\textit{FM})—a 500M-parameter model—that extracts concise entities from overly long responses when the final answer exceeds 100 characters. In such cases, \textit{FM} extracts the entity and returns it as the final answer.

\begin{algorithm}[t]
\small
\caption{Final Answer Decision}
\begin{algorithmic}
% \Require $sol_{cot}$, $A_{cot}$, $\mathcal{C}_{pot}$, $A_{pot}$, $\mathcal{C}_{sql}$, $A_{sql}$, Threshold ($\theta$)
% \midrule
\State C, P, TS $\gets$ \textit{CC}($T$, $Q$, $sol_{cot}$, $A_{cot}$, $\mathcal{C}_{pot}$, $A_{pot}$, $\mathcal{C}_{sql}$, $A_{sql}$)

\If {max(C, P, TS) $> \theta$}
    \State $A_f$ $\gets \arg\max_{A \in \{A_{cot}, A_{pot}, A_{sql}\}} \{$C, P, TS$\}$
\Else 
    \State $A_f$ $\gets$ \textit{JA}($T$, $Q$, $sol_{cot}$, $A_{cot}$, $\mathcal{C}_{pot}$, $A_{pot}$, $\mathcal{C}_{sql}$, $A_{sql}$)
\EndIf

\If {len($A_f$) $> 100$}
    \State $A_f$ $\gets$ \textit{FM}($A_f$)
\EndIf
\end{algorithmic}
\label{alg:finalans}
\end{algorithm}

\begin{table*}[t]
\centering
\small
\begin{tabular}{p{0.63cm}|p{2.4cm}||p{0.57cm}|p{0.57cm}|p{0.57cm}||p{0.57cm}|p{0.57cm}|p{0.57cm}||p{0.57cm}|p{0.57cm}|p{0.57cm}||p{0.57cm}|p{0.57cm}|p{0.57cm}} 
\toprule
\multicolumn{2}{c|}{ - } & \multicolumn{3}{c|}{\textit{TabLaP}} & \multicolumn{3}{c|}{\textit{SynTQA}} & \multicolumn{3}{c|}{\textit{MixSC}}& \multicolumn{3}{c}{\textbf{MATA}}\\

\midrule
 &
Models &
EM  &
fuzzy  &
F1  &
EM  &
fuzzy &
F1  &
EM &
fuzzy  &
F1  &
EM  &
fuzzy  &
F1  \\

\midrule
 &
llama3.2-3b &
0.188 &
0.290 &
0.247 &
\underline{0.597} &
\underline{0.654} &
\underline{0.602} &
0.201 &
0.303 &
0.252 &
\textbf{0.736} &
\textbf{0.766} &
\textbf{0.736} \\

 &
mistral-7b &
0.049 &
0.231 &
0.102 &
\underline{0.639} &
\underline{0.680} &
\underline{0.645} &
0.271 &
0.385 &
0.289 &
\textbf{0.861} &
\textbf{0.880} &
\textbf{0.861} \\

Small  &
phi4-mini-3.8b &
0.333 &
0.483 &
0.362 &
\underline{0.813} &
\underline{0.827} &
\underline{0.813} &
0.500 &
0.593 &
0.528 &
\textbf{0.819} &
\textbf{0.847} &
\textbf{0.819} \\

% Open &
% qwen2.5-coder-7b &
% 0.667 &
% 0.693 &
% 0.671 &
% \underline{0.910} &
% \underline{0.928} &
% \underline{0.910} &
% 0.681 &
% 0.732 &
% 0.690 &
% \textbf{0.924} &
% \textbf{0.932} &
% \textbf{0.924} \\

LLM &
qwen2.5-3b &
0.396 &
0.479 &
0.400 &
\underline{0.694} &
\underline{0.737} &
\underline{0.694} &
0.438 &
0.517 &
0.442 &
\textbf{0.868} &
\textbf{0.883} &
\textbf{0.868} \\

  &
qwen2.5-7b  &
0.444 &
0.522 &
0.444 &
\underline{0.813} &
\underline{0.866} &
\underline{0.815} &
0.597 &
0.657 &
0.597 &
\textbf{0.951} &
\textbf{0.955} &
\textbf{0.951} \\

%  &
% qwen2.5-14b  &
% 0.694 &
% 0.726 &
% 0.704 &
% \textbf{0.938} &
% \textbf{0.948} &
% \textbf{0.938} &
% 0.722 &
% 0.734 &
% 0.722 &
% \underline{0.910} &
% \underline{0.910} &
% \underline{0.910} \\
\midrule

 &
mistral-small-24b  &
0.764 &
0.784 &
0.773 &
\textbf{0.896} &
\textbf{0.918} &
\textbf{0.896} &
0.806 &
0.813 &
0.810 &
\textbf{0.896} &
\underline{0.896} &
\textbf{0.896} \\

 &
cogito-32b  &
\textbf{0.931} &
\textbf{0.934} &
\textbf{0.931} &
0.868 &
0.886 &
0.868 &
\underline{0.903} &
\underline{0.908} &
\underline{0.903} &
\underline{0.903} &
0.903 &
\underline{0.903} \\

Large &
qwen2.5-32b&
0.611 &
0.687 &
0.656 &
\underline{0.861} &
\underline{0.892} &
\underline{0.861} &
0.785 &
0.802 &
0.789 &
\textbf{0.917} &
\textbf{0.917} &
\textbf{0.917} \\

LLM &
GPT-4o&
0.653 &
0.655 &
0.653 &
\textbf{0.951} &
\textbf{0.961} &
\textbf{0.951} &
0.833 &
0.835 &
0.833 &
\underline{0.903} &
\underline{0.903} &
\underline{0.903} \\

 &
Claude-3.7-Sonnet&
0.868 &
0.868 &
0.868 &
\textbf{0.965} &
\textbf{0.970} &
\textbf{0.965} &
0.924 &
0.924 &
0.924 &
\underline{0.951} &
\underline{0.951} &
\underline{0.951} \\

\midrule
&
\textit{Average} &
0.524 &
0.593 &
0.544 &
\underline{0.810} &
\underline{0.839} &
\underline{0.811} &
0.626 &
0.674 &
0.637 &
\textbf{0.881} &
\textbf{0.890} &
\textbf{0.881} \\

\bottomrule
\end{tabular}
\caption{Evaluation results on the \textbf{Penguins in a Table} benchmark under our evaluation protocol. We report Exact Match(EM) accuracy, fuzzy matching, and F1 scores for each model. Bold indicates the best performance; underlined scores are the second best.}
\label{tab:penguins}
\end{table*}
\begin{table*}[t]
\centering
\small
\begin{tabular}{p{0.63cm}|p{2.4cm}||p{0.57cm}|p{0.57cm}|p{0.57cm}||p{0.57cm}|p{0.57cm}|p{0.57cm}||p{0.57cm}|p{0.57cm}|p{0.57cm}||p{0.57cm}|p{0.57cm}|p{0.57cm}} 
\toprule
\multicolumn{2}{c|}{ - } & \multicolumn{3}{c|}{\textit{TabLaP}} & \multicolumn{3}{c|}{\textit{SynTQA}} & \multicolumn{3}{c|}{\textit{MixSC}}& \multicolumn{3}{c}{\textbf{MATA}}\\

\midrule
 &
Models &
EM  &
fuzzy  &
F1  &
EM  &
fuzzy &
F1  &
EM &
fuzzy  &
F1  &
EM  &
fuzzy  &
F1  \\

\midrule
 &
llama3.2-3b &
0.067 &
0.357 &
0.130 &
\underline{0.089} &
0.231 &
0.120 &
0.081 &
\underline{0.372} &
\underline{0.144} &
\textbf{0.354} &
\textbf{0.563} &
\textbf{0.381} \\

&
mistral-7b &
0.036 &
0.331 &
0.119 &
\underline{0.227} &
\underline{0.367} &
\underline{0.270} &
0.082 &
0.355 &
0.151 &
\textbf{0.294} &
\textbf{0.473} &
\textbf{0.321} \\

Small &
phi4-mini-3.8b &
0.056 &
0.334 &
0.126 &
\underline{0.202} &
0.366 &
\underline{0.253} &
0.144 &
\underline{0.411} &
0.203 &
\textbf{0.273} &
\textbf{0.457} &
\textbf{0.295} \\

% &
% qwen2.5-coder-7b &
% 0.266 &
% 0.484 &
% 0.300 &
% \underline{0.374} &
% \underline{0.524} &
% \underline{0.412} &
% 0.254 &
% 0.487 &
% 0.286 &
% \textbf{0.444} &
% \textbf{0.625} &
% \textbf{0.479} \\

LLM  &
qwen2.5-3b &
0.163 &
\underline{0.417} &
0.195 &
\underline{0.208} &
0.364 &
\underline{0.245} &
0.163 &
\underline{0.417} &
0.197 &
\textbf{0.291} &
\textbf{0.471} &
\textbf{0.317} \\

  &
qwen2.5-7b  &
0.079 &
0.255 &
0.094 &
\underline{0.302} &
\underline{0.450} &
\underline{0.336} &
0.169 &
0.368 &
0.190 &
\textbf{0.354} &
\textbf{0.557} &
\textbf{0.393} \\

%  &
% qwen2.5-14b  &
% 0.325 &
% 0.583 &
% 0.372 &
% 0.341 &
% 0.489 &
% 0.377 &
% \underline{0.354} &
% \underline{0.592} &
% \underline{0.395} &
% \textbf{0.506} &
% \textbf{0.674} &
% \textbf{0.539} \\
\midrule

 &
mistral-small-24b  &
0.322 &
0.478 &
0.352 &
\underline{0.391} &
\underline{0.543} &
\underline{0.431} &
0.378 &
0.530 &
0.410 &
\textbf{0.573} &
\textbf{0.724} &
\textbf{0.606} \\

 &
cogito-32b  &
0.440 &
\underline{0.614} &
\underline{0.483} &
\underline{0.443} &
0.591 &
0.481 &
0.430 &
\underline{0.614} &
0.476 &
\textbf{0.577} &
\textbf{0.723} &
\textbf{0.609} \\

Large &
qwen2.5-32b&
0.268 &
0.533 &
0.317 &
\underline{0.398} &
\underline{0.553} &
\underline{0.436} &
0.297 &
0.551 &
0.341 &
\textbf{0.577} &
\textbf{0.721} &
\textbf{0.607} \\

LLM  &
GPT-4o&
\underline{0.556} &
\underline{0.722} &
\underline{0.595} &
0.476 &
0.607 &
0.503 &
0.494 &
0.692 &
0.540 &
\textbf{0.595} &
\textbf{0.740} &
\textbf{0.629} \\

 &
Claude-3.7-Sonnet&
0.612 &
0.763 &
0.655 &
0.489 &
0.633 &
0.540 &
\underline{0.619} &
\textbf{0.767} &
\underline{0.659} &
\textbf{0.620} &
\underline{0.764} &
\textbf{0.664} \\

\midrule
&
\textit{Average} &
0.260 &
0.480 &
0.307 &
\underline{0.322} &
0.471 &
\underline{0.362} &
0.286 &
\underline{0.508} &
0.331 &
\textbf{0.451} &
\textbf{0.619} &
\textbf{0.482} \\

\bottomrule
\end{tabular}
\caption{Evaluation results on the \textbf{TableBench} benchmark under our evaluation protocol. Bold and underline follow Table 2.}
\label{tab:tablebench}
\end{table*}

\section{Experiments}
\label{sec:mainexperiements}

\subsection{Baselines and Benchmarks}

To evaluate performance on open-source LLMs and new datasets not considered in prior research setups, we select three baselines whose official code is publicly available.

\noindent\textbf{SynTQA}~\cite{zhang-etal-2024-syntqa} ensembles text2SQL and end-to-end table QA (E2E TQA) models, leveraging their complementary strengths: text2SQL excels in numerical reasoning and handling large tables, while E2E TQA performs better with ambiguous questions. A lightweight selector (feature-based or LLM-based) chooses the final answer. For fair comparison, we used the same frozen LLM and prompts across all SynTQA components.

\noindent\textbf{MixSC}~\cite{Mix-SC} integrates textual and symbolic reasoning via a self-consistency mechanism. It utilizes GPT-3.5 for direct prompting and Python code execution, aggregating outputs for robustness. A normalization module (NORM) further enhances stability against structural perturbations.

\noindent\textbf{TabLaP}~\cite{TabLaP} employs multiple LLMs for table QA, delegating numerical reasoning to Python scripts generated by an LLM (NumSolver), while using MixSC \cite{Mix-SC} for non-numerical questions. An LLM-based selector (AnsSelector) chooses the more reliable answer, aided by a trustworthiness evaluator (TwEvaluator) for reliability estimation.

We exclude certain baselines from the main cross-backbone comparison for the following reasons: their official publicly released code is exclusively compatible with the closed-source GPT series LLMs \cite{chameleon, ReAcTable}, critical version dependencies that prevented execution \cite{Binder,DATER}, or significant performance degradation when switching to different models, rendering comparisons meaningless \cite{wang2024chainoftable, pal}. For GPT-specific tool-use baselines, we provide a separate GPT-4o-based comparison in Appendix~\ref{appendix:other_baselines}.

% \subsection{Benchmarks}

We used two benchmarks of varying difficulty. To verify that MATA does not rely on the distribution of the training datasets used for \textit{Sch} and \textit{CC}, we deliberately selected benchmark datasets different from the three datasets (WikiTQ, TabMWP, TabFact) used in training.

\noindent\textbf{Penguins in a Table}~\cite{penguins} is a diagnostic dataset from BIG-bench designed to test basic table reasoning. It presents a single table of penguin species with attributes such as height and weight, and asks simple factual or comparative questions. This isolates core table understanding without involving complex language or multi-step logic, making it a relatively easy task.

\noindent\textbf{TableBench}~\cite{tablebench} is a benchmark for complex table QA, spanning 18 subcategories including fact-checking, numerical reasoning, data analysis, and visualization. Tables are sourced from diverse domains such as finance, sports, and science. The benchmark emphasizes real-world complexity and demands diverse reasoning strategies.

\subsection{Metrics and Setup}

In many previous TableQA studies, Exact Match (EM) accuracy has been the most commonly used evaluation metric. However, EM alone is insufficient for assessing the quality of answers generated by LLMs.  Therefore, unlike prior work, we employed two additional metrics alongside EM. First, we used fuzzy matching\footnote{https://pypi.org/project/fuzzywuzzy/}, a metric widely adopted in studies \cite{king-flanigan-2024-unsupervised, cheng-etal-2024-dataflow, nekvinda-dusek-2021-shades} to measure textual similarity based on Levenshtein distance \cite{levenshtein1966binary}. Second, we adopted the SQuAD-style token-level F1 score \cite{rajpurkar-etal-2016-squad}, which evaluates token-level overlap between the prediction and the ground truth. By incorporating these two metrics, we complement the strictness of EM with more flexible and nuanced evaluations.

% %%%%%%%%% CR Add  %%%%%%%%%
% Specifically, \textbf{TableBench} contains heterogeneous target formats, including short entity or numeric answers, sentence-level descriptive outputs, and chart-generation tasks. Since our focus is entity-type TableQA, where the expected output is a concise entity or value, we evaluate the entity-answer subset of TableBench (693 out of 886 examples) and use EM, fuzzy matching, and token-level F1 as the main metrics. We exclude examples whose targets are not short entity-style answers, such as those in the Descriptive Analysis, Anomaly Detection, Causal Analysis, and Chart Generation categories. To improve comparability with the original \textbf{TableBench} evaluation, we additionally report ROUGE-L results on the same subset in Appendix~\ref{app:rouge_l}.
% %%%%%%%%% CR Add  %%%%%%%%%

% \subsection{Setup}
The MATA framework is implemented using LangChain\footnote{https://www.langchain.com/} and leveraged Ollama\footnote{https://ollama.com/} to support open-source LLMs. This setup enables flexible model switching, provided that the models are supported by Ollama. We defined LLMs with parameter sizes under 10B as \textit{small LLMs} \cite{meta-llama‑3.2-blog, jiang2023mistral7b, qwen2025qwen25technicalreport}, and those with more than 10B parameters or that are closed-source as \textit{large LLMs} \cite{mistral_small_3_2025, deepcogito_cogito_v1_2025, qwen2025qwen25technicalreport, openai2024gpt4ocard, anthropic_claude37_blog2025}. We used five models per category, totaling ten models across both small and large LLMs. MATA and all locally executable baseline runs were evaluated under the same inference conditions using a single A100 GPU.

\subsection{Results}

On the easier benchmark \textbf{Penguins in a Table} (Table~\ref{tab:penguins}), MATA maintained consistently high performance across all 10 models, clearly demonstrating its strong model-agnostic capability. In contrast, despite the simplicity of the tables and questions, the baselines TabLaP and MixSC exhibited significant performance declines when applied to smaller LLMs. This indicates that smaller LLMs possess insufficient capabilities for table analysis, and approaches such as TabLaP and MixSC fail to adequately address this shortfall. 

While SynTQA achieved highest performance among baselines for some small LLMs on this benchmark, this advantage was confined to simple, single-table reasoning tasks. We attribute this to its avoidance of excessive reasoning. Whereas TabLaP and MixSC entail more than ten LLM inferences to derive a final answer, SynTQA requires only three. This finding aligns with prior findings \cite{huang2025is, NEURIPS2024_51173cf3}, which posits that repeated inference in straightforward tasks can lead to performance degradation. 

However, this design choice comes at the expense of generalizability and flexibility in more complex tasks. On \textbf{TableBench} (Table~\ref{tab:tablebench}), which includes larger tables and more challenging questions, MATA achieved the best performance across all models. For example, MATA outperformed the best baseline (SynTQA) by 40.1\% (EM), 21.9\% (fuzzy), and 33.1\% (F1). Notably, SynTQA did not maintain its lead on these harder tasks, and in some cases was outperformed by TabLaP and MixSC, especially when leveraging closed-source models. This suggests that approaches relying on minimal reasoning steps, such as SynTQA, may not generalize well to complex table understanding scenarios, where a more nuanced and adaptive reasoning strategy is necessary. Furthermore, TabLaP and MixSC, originally evaluated only with closed-source GPT series models, are specifically tuned for such environments and do not transfer robustly to open-source or smaller LLMs. In contrast, MATA employs multiple types of reasoning only to the extent necessary, achieving optimal performance across various models. In other words, MATA demonstrates superior generalization and stability, delivering optimal performance across a diverse range of models and task difficulties. This underscores the practical value of MATA's balanced and flexible reasoning framework for diverse TableQA applications, where adaptability and reliability are crucial.

%%%%%%%%% CR Add  %%%%%%%%%
We also compare MATA with additional TableQA and multi-agent baselines in Appendix~\ref{appendix:other_baselines}. With GPT-4o as the shared backbone, MATA substantially outperforms ReAcTable \cite{ReAcTable} and Chameleon \cite{chameleon} on both \textbf{Penguins in a Table} and \textbf{TableBench}. We further compare MATA against AutoPrep \cite{AutoPrep}, another standard multi-agent framework, in which a single LLM backbone is tasked with planning, executing, and iteratively refining answers for question-aware data preparation in TableQA, using representative Qwen2.5 backbones. MATA performs better in most settings, suggesting that its gains are not limited to the three main baselines in Tables~\ref{tab:penguins} and~\ref{tab:tablebench}.

Furthermore, since the original \textbf{TableBench} study evaluates performance using the ROUGE-L metric, we provide corresponding ROUGE-L scores in Appendix~\ref{app:rouge_l} to facilitate direct comparison.
%%%%%%%%% CR Add  %%%%%%%%%

\begin{figure}[t!]
\centering
\includegraphics[width=\linewidth]{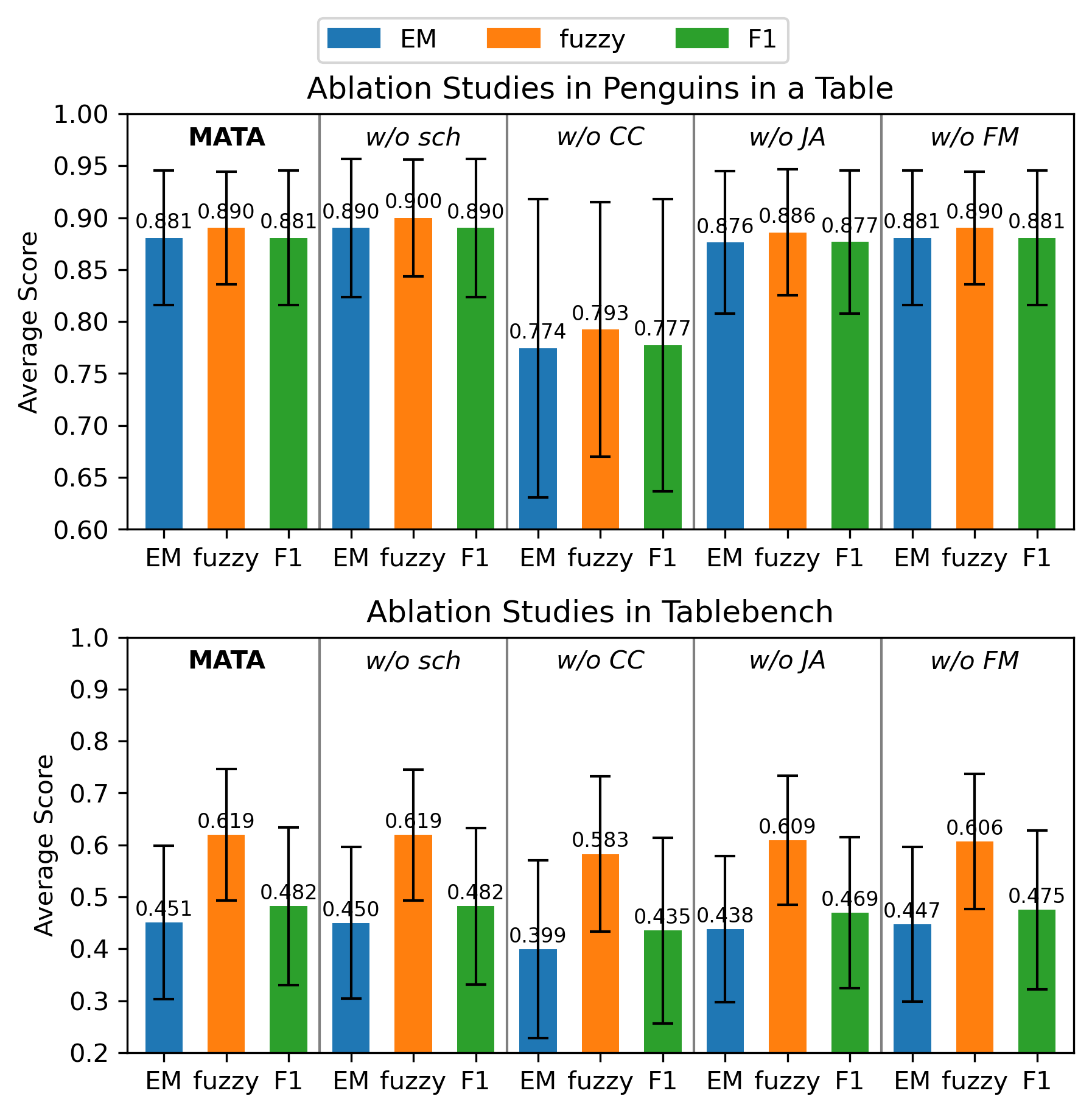} 
\caption{Ablation Studies on the \textbf{Penguins in a Table} (top) and \textbf{TableBench} (bottom). The scores shown in the graph represent the average across all models.}
\label{fig:ablation}
\end{figure}

\subsection{Ablation Study}

We also conduct an ablation study to analyze the contribution of each module in MATA. Figure~\ref{fig:ablation} reveals that the Confidence Checker (\textit{CC}) is the most critical component for overall performance.
Removing \textit{CC} from the framework results in the largest performance drop in accuracy, as it forces the system to rely solely on the Judge Agent (\textit{JA}) to evaluate all candidate answers, which is a much less efficient strategy.

The main strength of the \textit{CC} lies in its ability to bypass unnecessary reasoning when a candidate's confidence score is sufficiently high, allowing immediate selection of that answer without invoking additional LLM-based inference through the \textit{JA}. This not only improves computational efficiency but also helps avoid performance degradation caused by excessive inference steps or prompts, a phenomenon reported in prior studies~\cite{huang2025is, NEURIPS2024_51173cf3}.

Empirical results demonstrate that using \textit{CC} significantly reduced the \textit{JA} invocation frequency by 95.8\% on the easier \textbf{Penguins in a Table} and by 60.6\% on the more challenging \textbf{TableBench}, all while maintaining or improving final accuracy (see Appendix~\ref{appendix:F} for details). This demonstrates that efficient candidate selection, rather than exhaustive verification, yields both better performance and greater efficiency. The roles of the Judge Agent (\textit{JA}) and the Format Matcher (\textit{FM}) become more significant as task complexity increases. On simple benchmarks like \textbf{Penguins in a Table}, most correct answers can be identified directly by the \textit{CC}, with little need for further processing. However, on challenging tasks such as \textbf{TableBench}, the additional verification and answer formatting provided by \textit{JA} and \textit{FM} consistently result in superior outcomes across all models. This pattern illustrates that modular, adaptive answer selection is especially important for complex TableQA tasks.

We also evaluated the impact of the Scheduler (\textit{Sch}) module. Using the \textit{Sch} led to a 14.6\% reduction in LLM agent calls on \textbf{Penguins in a Table} and a 7.6\% reduction on \textbf{TableBench}, aggregated across all 10 models, further improving computational efficiency (see Appendix~\ref{appendix:F} for details). However, its contribution to overall accuracy was context-dependent. In high-difficulty settings (e.g., \textbf{TableBench}), omitting the \textit{Sch} and using all available reasoning paths sometimes led to marginally better results, likely due to increased answer diversity. Conversely, in easier tasks (e.g., \textbf{Penguins in a Table}), reducing the number of reasoning methods occasionally limited answer diversity, slightly diminishing confidence in the final prediction.

These results suggest a nuanced trade-off between efficiency and answer diversity. For complex table understanding problems, using all reasoning paths can, in some cases, increase the risk of spurious or conflicting inferences, potentially hindering the selection of the correct answer. In such scenarios, a moderate and well-chosen degree of reasoning (i.e., not simply maximizing the number of paths) tends to yield the best balance of efficiency and accuracy. Similarly, for low-difficulty tasks involving small tables and straightforward questions, we observed that eliminating certain reasoning methods by using the \textit{Sch} sometimes led to a slight decrease in answer diversity, and thus in confidence or robustness of the final prediction. This underscores the value of generating a sufficiently diverse set of candidate answers, even in seemingly simple settings.

Therefore, our findings indicate that achieving optimal TableQA performance requires a careful balance between ensuring sufficient reasoning diversity and optimizing the number of inference steps. Strategic use of the \textit{Sch} and other modules is essential: too many reasoning paths may introduce noise, while too few may miss important perspectives. These observations reinforce the modular design philosophy of MATA, highlighting that both efficiency and robustness are best served by flexible, context-aware inference orchestration.

\subsection{End-to-End Efficiency}
\label{sec:e2e_efficiency}

\begin{table}[h]
\centering
\small
\begin{tabular}{p{2.25cm}p{0.8cm}p{0.95cm}p{0.75cm}p{0.7cm}}
\toprule
Model & TabLaP & SynTQA & MixSC & \textbf{MATA} \\
\midrule
llama3.2-3b & 15.44 & 5.95 & 12.58 & 21.32 \\
mistral-7b & 23.80 & 3.34 & 20.69 & 26.63 \\
phi4-mini-3.8b & 12.49 & 10.42 & 9.66 & 23.26 \\
qwen2.5-3b & 20.47 & 0.93 & 18.13 & 17.77 \\
qwen2.5-7b & 18.83 & 1.65 & 16.80 & 20.82 \\
mistral-small-24b & 93.88 & 12.45 & 84.07 & 47.31 \\
cogito-32b & 127.84 & 14.75 & 121.30 & 31.83 \\
qwen2.5-32b & 78.35 & 5.36 & 72.61 & 31.48 \\
\midrule
Average & 48.89 & 6.86 & 44.48 & 27.55 \\
\bottomrule
\end{tabular}
\caption{Average end-to-end latency in seconds per query for locally hosted open-source backbones. Latency is measured in the same local environment and averaged over \textbf{Penguins in a Table} and \textbf{TableBench}. Lower is better.}
\label{tab:e2e_latency}
\end{table}

We further evaluate the end-to-end efficiency of MATA by measuring wall-clock response latency. Since closed-source API models introduce provider-side scheduling, network, and service-load overheads that are not directly comparable to local inference, we report latency only for open-source backbones executed in the same local environment. 

Table~\ref{tab:e2e_latency} reports the average end-to-end latency per query. In the same local environment, the combined latency of the lightweight \textit{Sch}/\textit{CC}/\textit{FM} tools accounts for only about 0.6\% of MATA’s total latency, confirming that the dominant cost comes from backbone LLM invocations. SynTQA is the fastest method because it uses a fixed inference budget with minimal LLM calls (only 3). However, this also limits its flexibility when handling more complex questions, resulting in lower performance, as shown in Table~\ref{tab:tablebench}. In contrast, TabLaP and MixSC require heavier fixed inference budgets (12 and 10 LLM calls, respectively) and therefore incur substantially higher latency. MATA lies between these two extremes: it allocates additional computation when needed, but avoids unnecessary reasoning branches and Judge Agent (\textit{JA}) calls through \textit{Sch}- and \textit{CC}-driven early exits. As a result, MATA achieves lower average latency than the call-heavy baselines while maintaining the strongest overall accuracy, especially on the more challenging \textbf{TableBench}.

\section{Conclusion}
%조성환 conclusion 수정본
We introduce MATA, a novel multi-agent framework for reliable and flexible Table Question Answering, which leverages diverse reasoning paths including Chain-of-Thought, Program-of-Thought, and text-to-SQL to generate multiple candidate answers, while employing lightweight tools and specialized agents to optimize selection and minimize costly LLM inferences. Overall, MATA advances TableQA capabilities by highlighting model-agnostic design and efficient reasoning with future work potentially integrating additional techniques or agents to broaden its applicability across diverse scenarios and data complexities.

% In this paper, we introduce MATA, a multi-agent framework for reliable TableQA that orchestrates diverse reasoning paths, including CoT, PoT, and text-to-SQL. By employing lightweight tools to optimize answer selection, MATA achieves high accuracy while significantly minimizing costly LLM inferences. This model-agnostic approach demonstrates the importance of efficient reasoning, with future work aiming to broaden its applicability across diverse data complexities.

\section*{Limitations}

While MATA reduces inference cost by selectively executing reasoning methods and utilizing lightweight modules such as the scheduler and selector, it does not address the fundamental cost associated with the LLMs themselves. Each reasoning path still requires full LLM inference, which can be computationally expensive, especially for large-scale models. As a result, the overall efficiency of the system remains constrained by the inherent resource demands of LLM-based reasoning. Future work should explore LLM compression, distillation, or hybrid architectures that offload parts of the reasoning process to smaller or non-LLM components.

\section*{Ethical Considerations}

Our proposed MATA framework, while designed to improve the reliability and efficiency of TableQA systems, can potentially be misused in ways that raise ethical concerns. Specifically, the model’s ability to generate seemingly plausible answers via multiple reasoning paths (e.g., CoT, PoT, text2SQL) may be exploited to fabricate misleading tabular information or circumvent truthfulness checks in sensitive applications such as finance, healthcare, or public policy. Additionally, our model inherits biases from the underlying LLMs and datasets, which could manifest as systematic errors or unfair predictions, particularly when deployed without careful calibration. Although MATA is designed to work with open-source LLMs to encourage accessibility and transparency, this flexibility may also increase the risk of improper use in low-accountability environments. We strongly discourage the application of MATA in domains requiring verifiable factuality and fairness without additional safeguards. We advocate for its use in responsible, research-driven, and low-risk settings and encourage further research into robust hallucination detection and bias mitigation to enhance the ethical deployment of TableQA models. In accordance with the ACL Policy on AI Assistance, we acknowledge the use of Gemini\footnote{\url{https://deepmind.google/technologies/gemini/}} to assist with code debugging and writing polishing. All experimental designs, data analyses, and scientific claims presented in this work were verified by the authors.

\bibliography{custom}

\newpage

\appendix

% \onecolumn

\section*{Appendix}

\section{Model Performance Differences Depending On The Reasoning Path}
\label{appendix:A}

\begin{figure*}[h!]
  \makebox[\linewidth][c]{
    \includegraphics[width=5.5in]{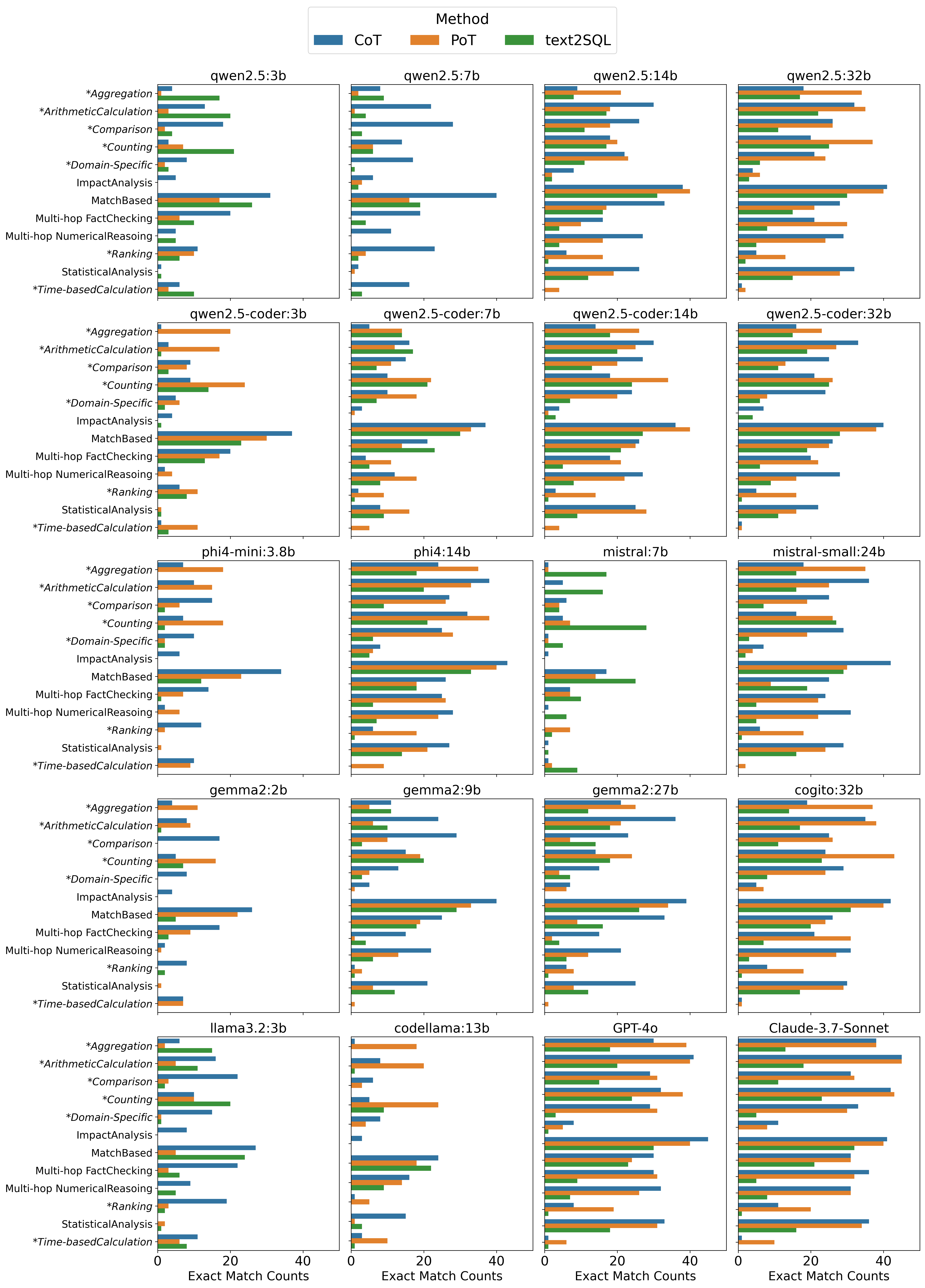}
  }
  \caption{Exact Match (EM) accuracy comparison of CoT, PoT, and text2SQL across LLMs on the TableBench \cite{tablebench} dataset. The figure highlights each method’s strengths and weaknesses by question category. Asterisks on the y-axis indicates categories related to Numerical Reasoning.}
\label{fig:appendixA}
\end{figure*}

We analyze the relative strengths of three major reasoning strategies for Table QA—Chain-of-Thought (CoT) \cite{CoT}, Program-of-Thought (PoT) \cite{PoT}, and text-to-SQL (Text2SQL) \cite{zhong2017seq2sqlgeneratingstructuredqueries}—across different reasoning categories. In total, we evaluate 20 models, including 18 open-source models ranging in size from 3B to 32B and two closed-source models (See Figure~\ref{fig:appendixA}). For a fair comparison, we use only the prompts defined in Appendix~\ref{appendix:sectionG} for the CoT Agent (CoTA), PoT Agent (PoTA), and Text2SQL Agent (t2SA), excluding Debug Agents that might bias results in favor of CoTA.

To evaluate under realistic and challenging conditions, we conduct experiments on TableBench \cite{tablebench}, a recently introduced benchmark that spans 18 subcategories across four reasoning domains, including numerical reasoning and fact checking. TableBench provides a more faithful reflection of real-world scenarios and enables fine-grained performance analysis across diverse question types.

As shown in Figure~\ref{fig:appendixA}, in several models,  and consistent with prior findings \cite{Mix-SC, zhang-etal-2024-syntqa, TabLaP}, CoT tends to perform well on fact-checking and data interpretation tasks, while PoT and text2SQL show better results for complex arithmetic and aggregation-based questions. This trend is particularly evident in both closed-source models. That is, the patterns observed in prior studies appear more frequently in models with 10B+ parameters or closed-source configurations.

However, our results also reveal that such trends vary substantially depending on the model series and size. For instance, among 7B-sized models, text2SQL excels in mistral, PoT dominates in qwen2.5-coder, and CoT performs best in qwen2.5. Within the qwen2.5 family, smaller models like 3B and 7B favor CoT the most, followed by text2SQL, with PoT lagging behind. As the size increases to 14B and 32B, PoT’s performance improves dramatically. In the code-specialized qwen2.5-coder series, PoT significantly outperforms its performance in the general-purpose qwen2.5 models.

%In conclusion, there is no single dominant reasoning strategy across model sizes, series, or access types (open vs. closed source). Numerous exceptions arise depending on the model, making it difficult to establish consistent trends. These findings suggest that the effectiveness of each reasoning strategy depends not only on the question type but also heavily on model capacity and specialization. While prior works have broadly categorized reasoning methods by task type, our analysis shows that such classifications do not generalize across models of different scales and series. Therefore, flexible and adaptive method selection is essential for robust and reliable Table QA systems.

In conclusion, we find no consistent pattern in the performance of CoT, PoT, or text2SQL across factors such as model size, architecture series, or whether the model is open- or closed-source. Numerous exceptions emerge depending on the specific model, making it difficult to draw universal trends. These findings suggest that the effectiveness of a reasoning strategy is influenced not only by the question type but also by a wide range of other factors, including model capacity and internal design. While previous studies \cite{Mix-SC, zhang-etal-2024-syntqa, TabLaP} typically categorized reasoning methods based on task types, our analysis shows that such categorizations do not generalize well across different model scales or series. Therefore, robust Table QA requires flexible and dynamic selection of reasoning strategies.

\section{Details of Training Dataset, Scheduler(\textit{sch}) and Confidence Checker(\textit{CC})}
\label{appendix:B}

This section provides detailed information about the training datasets as well as the specific implementation of the scheduler (\textit{sch}) and the Confidence Checker (\textit{CC}).

\subsection{Training Dataset for \textit{Sch} and \textit{CC}} 
\label{sec:training}

We construct a large-scale training dataset specifically designed to support the fine-tuning of lightweight tools—\textit{Sch} and \textit{CC}—within the multi-reasoning TableQA framework MATA.

To build this dataset, we leverage three publicly available TableQA datasets: WikiTQ \cite{pasupat-liang-2015-compositional}, TabMWP \cite{lu2023dynamic}, and TabFact \cite{Chen2020TabFact}. Detailed statistics for each source dataset are provided in Table~\ref{tab:dataset-summary}.

% \begin{table}[h]
% \centering
% \scalebox{0.85}{
% \begin{tabular}{l|c|c|c|c|c|c}
% \toprule
% \textbf{Dataset} & \textbf{\#Train} & \textbf{\#Val} & \textbf{\#Test} & \textbf{\# of table cells} & \textbf{Main Task} & \textbf{Table Source} \\
% \midrule
% WikiTQ & 11,321 & 2,831 & 4,344 & 162.3 & Compositional QA & Wikipedia (HTML tables) \\
% TabMWP & 23,059 & 7,686 & 7,686 & 11.3 & Multi-step Math QA & Curated semi-structured tables \\
% TabFact & 92,283 & 12,792 & 12,779 & 85.6 & Fact Verification & Wikipedia (infobox-style tables) \\
% \bottomrule
% \end{tabular}
% }
% \caption{Summary of table QA datasets used for training dataset.}
% \label{tab:dataset-summary}
% \end{table}

\begin{table*}[t!]
\centering
\scalebox{0.85}{
\begin{tabular}{l|c|c|c|c|c}
\toprule
\textbf{Dataset} & \textbf{\#Train} & \textbf{\#Val} & \textbf{\#Test}  & \textbf{Main Task} & \textbf{Table Source} \\
\midrule
WikiTQ & 11,321 & 2,831 & 4,344  & Compositional QA & Wikipedia (HTML tables) \\
TabMWP & 23,059 & 7,686 & 7,686  & Multi-step Math QA & Curated semi-structured tables \\
TabFact & 92,283 & 12,792 & 12,779 & Fact Verification & Wikipedia (infobox-style tables) \\
\bottomrule
\end{tabular}
}
\caption{Summary of table QA datasets used for training \textit{CC} and \textit{sch}.}
\label{tab:dataset-summary}
\end{table*}

\noindent\textbf{\textbullet{} WikiTQ}~\cite{pasupat-liang-2015-compositional} is a widely used benchmark for table-based question answering, consisting of natural language questions over semi-structured Wikipedia tables. Each question is paired with a crowd-annotated answer. The dataset emphasizes compositional reasoning, covering operations such as superlatives, aggregation, and arithmetic. Tables are disjoint across splits to test generalization to unseen schemas.

\noindent\textbf{\textbullet{} TabMWP}~\cite{lu2023dynamic} is a dataset of math word problems grounded in semi-structured tables, requiring multi-step numerical reasoning. Questions are mostly free-text with diverse answer types including integers, decimals, and spans. It combines textual and tabular cues, posing challenges in alignment and multi-hop symbolic reasoning.

\noindent\textbf{\textbullet{} TabFact}~\cite{Chen2020TabFact} is a large-scale fact verification dataset containing human-annotated statements labeled as ENTAILED or REFUTED, grounded in Wikipedia tables. It requires both linguistic inference and symbolic reasoning. Statements vary in complexity, from simple row-level facts to multi-step logical compositions involving comparison and aggregation.

Specifically, we use the entire training and validation sets from WikiTQ (14,152 examples), examples with indices 0 through 23,006 from the TabMWP training set (23,007 examples), and examples with indices 0 through 20,728 from the TabFact training set (20,729 examples).

A total of 57,888 unique table-question pairs (\textit{T}, \textit{Q}) are processed using three large language models— CodeLLaMA:13B \cite{codellama}, phi4:14B \cite{phi4technicalreport}, and Qwen2.5-Coder:14B \cite{qwen25codertechnicalreport}—to yield 173,664 samples. For each (\textit{T}, \textit{Q}) pair, we generate outputs from three reasoning paradigms: Chain-of-Thought (CoT), Program-of-Thought (PoT), and text2SQL, all executed using the same inference pipeline as in the MATA framework. Specifically, CoT reasoning is performed once using the CoT Agent (CoTA), while PoT and text2SQL utilize inference-and-debug loops via the PoT Agent (PoTA) and Python Debug Agent (PDA), and the text2SQL Agent (t2SA) and SQL Debug Agent (SDA), respectively. (For further details on the reasoning and interactions among LLM agents, refer to the Methodology section of the main paper.)

\begin{figure*}[h!]
\centering
\includegraphics[width=0.96\textwidth]{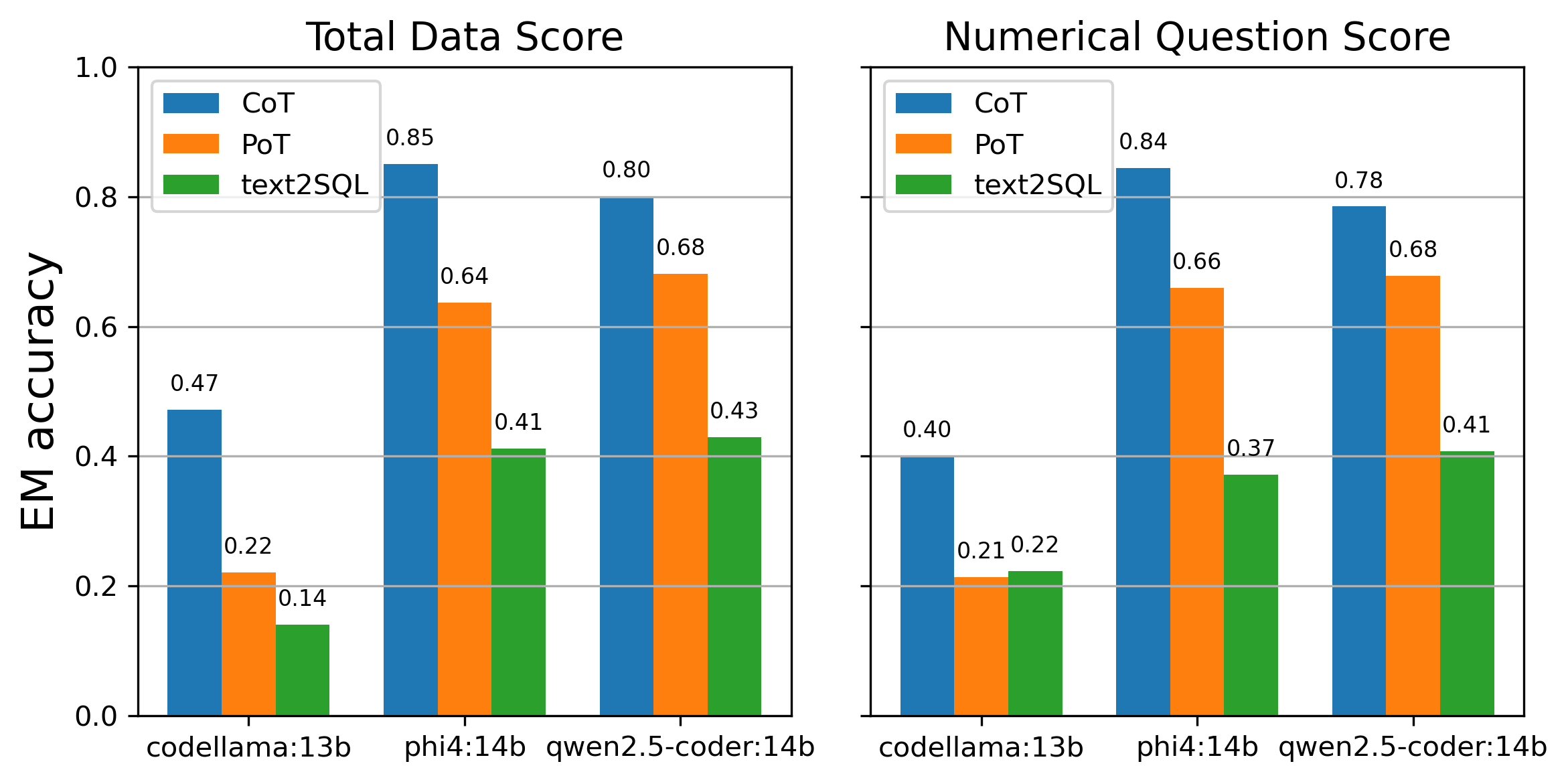} 
\caption{Exact Match (EM) accuracy on total training datasets: overall accuracy (left) and accuracy on numerical questions only (right). The x-axis represents different LLMs.}
\label{fig:traindata}
\end{figure*}

To foster research on improving the efficiency and adaptability of TableQA systems, the dataset is released at \url{https://github.com/AIDASLab/MATA/tree/main/experiments} for free academic use—along with documentation and licensing. For detailed statistics and structure of the training dataset, see Table~\ref{tab:train-dataset-summary} and Figure~\ref{fig:traindata}.

\begin{table*}[t!]
\centering
\scalebox{0.80}{
\begin{tabular}{c|c|c||c|c|c|c|c}
\toprule
\textbf{Dataset} & \textbf{\# (\textit{T}, \textit{Q}) pairs} & \textbf{\# table cells} & \textbf{LLM} & \textbf{\# CoT} & \textbf{\# PoT}  & \textbf{\# text2SQL} & \textbf{\# Incorrect All} \\
\midrule
       &        &       & CodeLLaMA:13B      & 4,832 & 3,813 & 3,284 & 6,378 \\
WikiTQ & 14,152 & 162.3 & phi4:14B           & 10,034 & 7,877 & 5,117 & 2,578\\
       &        &       & Qwen2.5-Coder:14B  & 8,802 & 8,290 & 5,625 & 2,795\\
 \midrule
       &        &       & CodeLLaMA:13B      & 11,403 & 2,988 & 4,542 & 9,326\\
TabMWP & 23,007 & 11.3  & phi4:14B           & 21,592 & 16,806 & 8,464 & 1,004 \\
       &        &       & Qwen2.5-Coder:14B  & 21,254 & 17,536 & 9,055 & 1,065\\
 \midrule
        &        &       & CodeLLaMA:13B      & 11,044 & 5,961 & 289 & 7,638\\
TabFact & 20,729 & 85.6  & phi4:14B           & 17,580 & 12,188 & 10,249 & 1,655\\
        &        &       & Qwen2.5-Coder:14B  & 16,313 & 13,601 & 10,173 & 1,863\\
\bottomrule
\end{tabular}
}
\caption{Summary of table QA datasets used for the training \textit{CC} and \textit{sch}.
\textbf{\# table cells} denotes the average number of cells per table in each dataset.\textbf{\# CoT}, \textbf{\# PoT}, and \textbf{\# text2SQL} indicate the number of questions correctly answered by each reasoning method according to the Exact Match metric.}
\label{tab:train-dataset-summary}
\end{table*}

\subsection{Scheduler (\textit{Sch})}

MATA optionally includes a scheduler module to optimize inference efficiency. The scheduler predicts which code-based reasoning path—PoT or text2SQL—is more likely to succeed, allowing MATA to prioritize that path while deferring or skipping the other.

The scheduler (denoted as \textit{Sch}) serves as the initial module, processing the input table (\textit{T}) and question (\textit{Q}). Based on various features extracted from \textit{T} (e.g., table size, schema, data types) and the semantic content of \textit{Q}, \textit{Sch} determines whether to perform PoT or text2SQL reasoning first. It is implemented as a lightweight model combining MobileBERT \cite{mobilebert} and a two-layer MLP, totaling 24.65M parameters.

\begin{figure}[h]
\centering
\includegraphics[width=0.5\textwidth]{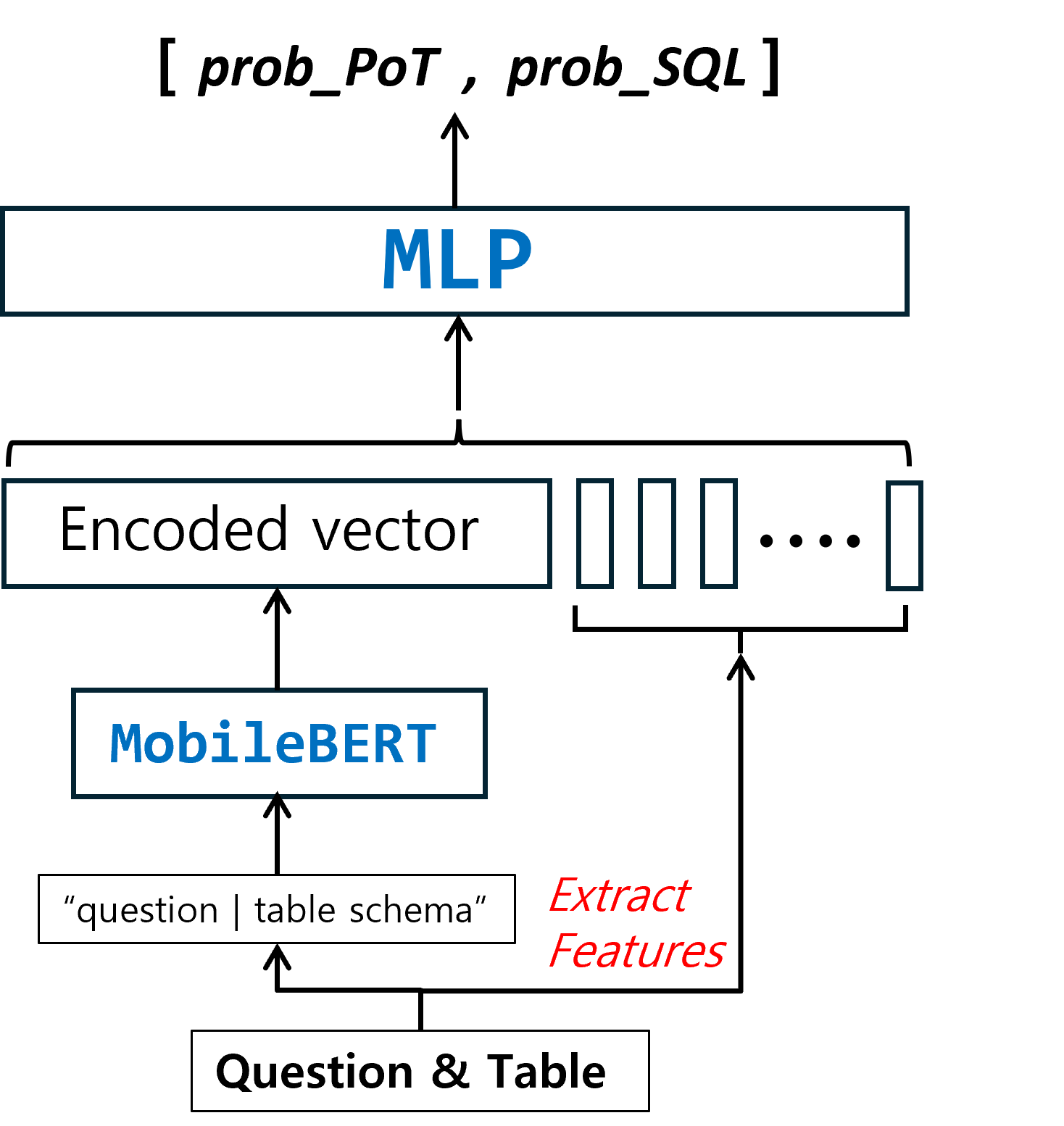} 
\caption{The architecture of the scheduler module in MATA. The scheduler encodes the input question and table schema using MobileBERT, and then concatenates the resulting vector with ten hand-crafted features extracted from the table and question. A two-layer MLP processes this combined representation and outputs the probabilities of success for the PoT and text2SQL reasoning paths, allowing MATA to prioritize the more promising one during inference.}
\label{fig:sch}
\end{figure}

The architecture of the scheduler (\textit{Sch}) is illustrated in Figure~\ref{fig:sch}. The scheduler first uses MobileBERT \cite{mobilebert}, a lightweight text encoder, to embed a concatenated text input that includes both the question and the table schema. Additional hand-crafted features are then concatenated to this embedded vector. These features are used as cheap routing cues for deciding whether PoT or text2SQL should be attempted first. Row, column, and cell counts capture table scale, where symbolic querying can be more useful for large tables. Question length and numeric-token counts capture question complexity and arithmetic demand. Schema-question overlap reflects column-localization ambiguity, and value-type flags indicate whether numerical computation, string matching, or missing-value handling is likely to dominate. These features are used only as inputs to \textit{Sch}, not as direct decision rules. This design is motivated by SynTQA’s feature-based selector and MixSC’s error analysis, which identify table scale, numeric demand, and schema-question localization as important cues for choosing reliable symbolic reasoning paths.

% These features include statistical and semantic information such as table size, lexical overlap between the question and schema, and the distribution of value types within the table cells. Specifically, the following ten features are appended as numeric or boolean values:

\paragraph{\textbf{Numeric features (6):}}
\begin{itemize}
\item Number of rows in the table
\item Number of columns in the table
\item Overall table size
\item Number of unique words in the question
\item Number of numeric tokens in the question
\item Number of overlapping words between the question and the table schema
\end{itemize}

\paragraph{\textbf{Boolean features (4):}}
\begin{itemize}
\item Whether integer-type values exist in table cells
\item Whether float-type values exist
\item Whether string-type values exist
\item Whether NaN values exist
\end{itemize}

In total, ten additional features are appended to the embedding vector produced by MobileBERT. This combined vector is passed through a two-layer Multi-Layer Perceptron (MLP), which outputs a final one-dimensional vector of length 2.

Training was conducted using the dataset described in Appendix~\ref{sec:training}. Specifically, for each Table-Question pair (\textit{T}, \textit{Q}), we label whether the PoT and text2SQL reasoning methods produce the correct answer based on the Exact Match metric, assigning binary values (0 or 1). These labels are then used to construct two-dimensional ground truth vectors, which supervise the fine-tuning of the scheduler (\textit{Sch}). The model is trained with Binary Cross Entropy loss for 45 epochs, using a batch size of 128 on a single NVIDIA A100 GPU.

At inference time, the scheduler outputs two logits, and MATA executes the reasoning path corresponding to the higher logit first, alongside CoT (Chain-of-Thought) reasoning. If the two results agree, the remaining reasoning path is skipped. If they disagree, the third path is also executed, and all three outputs are passed to the final answer selection stage. In this way, the scheduler reduces unnecessary LLM calls when confident, while preserving robustness by ensuring full fallback execution when disagreement occurs. When computational cost is not a concern, the scheduler can be disabled entirely to allow all three reasoning paths to run by default.

% \newpage

\subsection{Confidence Checker (\textit{CC})} 

The Confidence Checker (hereafter \textit{CC}) module takes as input the table \textit{T}, question \textit{Q}, and candidate answers generated from CoT, PoT, and text2SQL reasoning methods, assigning a confidence score to each candidate. Prior research \cite{gu-etal-2022-pasta} has reported strong performance of DeBERTaV3 \cite{he2023debertav3} on table understanding tasks. Consequently, \textit{CC} was implemented by fine-tuning DeBERTaV3-large, which has 435M parameters.

Key techniques used in implementing the Confidence Checker are as follows:

\subsubsection{Soft Labeling}

\begin{figure}[h]
\centering
\includegraphics[width=0.5\textwidth]{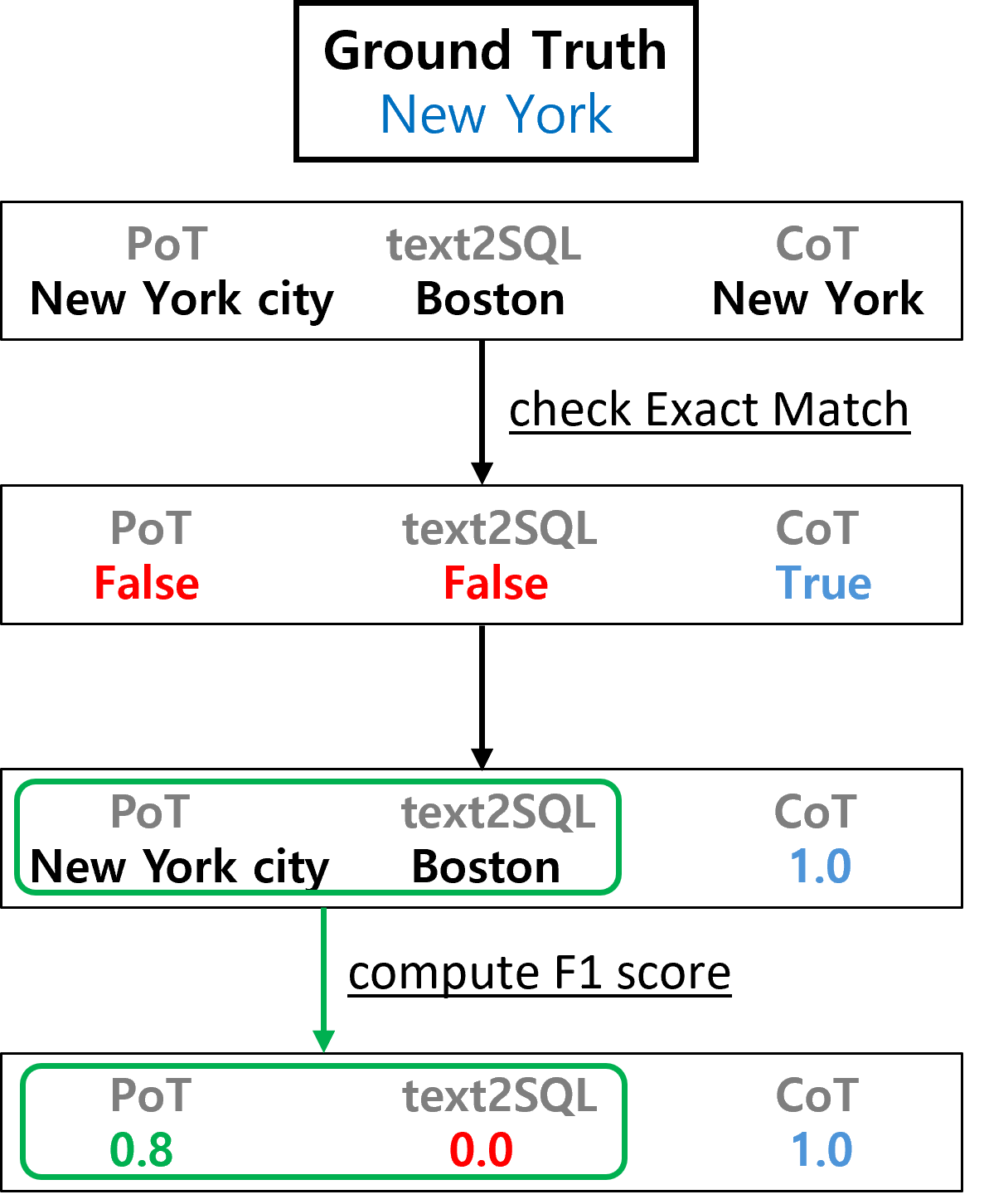} 
\caption{This figure shows the labeling rules for training the Confidence Checker. Soft labeling is performed in two orders: (1) If the final answer obtained from a method exactly matches the ground truth, it is labeled as 1.0. (2) When exact matching fails, we assign the label based on the F1 score \cite{rajpurkar-etal-2016-squad} between the model’s answer and the ground truth. In the case of CoT, exact matching is successful and thus directly used for supervision.}
\label{fig:CC}
\end{figure}

Initially, the CC is fine-tuned on a previously collected training dataset, taking each reasoning method's textual outputs as input and producing a 1-dimensional vector of length 3. Each element of this vector represents the confidence score for the respective PoT, text2SQL, and CoT reasoning paths. To perform supervised fine-tuning of DeBERTaV3-large, we labeled each vector by comparing the final predictions of each reasoning method with the ground truth. Rather than simply labeling using binary results (0 or 1) based on Exact Match, we employed a two-stage labeling process.
Figure~\ref{fig:CC} illustrates this procedure. First, predictions are evaluated using Exact Match; predictions with Exact Match results labeled as True are assigned a score of 1.0. Subsequently, for predictions that failed the Exact Match, an additional partial score was assigned based on F1-score evaluation. This allowed for semantic diversity in labeling, even if predictions did not exactly match the ground truth.

\subsubsection{Special Tokens}

XML-style structured special tokens were introduced to stabilize the training of DeBERTaV3-large. The added tokens are listed in Table~\ref{tab:special_tokens}.

\begin{table*}[t!]
\centering
% \resizebox{\columnwidth}{!}{%  <-- table* 환경에서는 전체 너비를 쓰므로 resizebox는 보통 뺍니다.
% \small % 내용이 많으므로 폰트 크기를 약간 줄이는 것을 추천합니다.
\begin{tabular}{l|p{13cm}} % p{너비}를 사용하여 내용이 길어지면 자동 줄바꿈되게 설정
\hline
\textbf{Category} & \textbf{Special Tokens} \\ \hline
Table Metadata & \texttt{<Table\_row\_size>}, \texttt{</Table\_row\_size>}, \texttt{<Table\_column\_size>}, \texttt{</Table\_column\_size>}, \texttt{<Table\_size>}, \texttt{</Table\_size>}, \texttt{<Table>}, \texttt{</Table>} \\
Input/Output & \texttt{<Question>}, \texttt{</Question>}, \texttt{<solution>}, \texttt{</solution>}, \texttt{<answer>}, \texttt{</answer>} \\
Reasoning Types & \texttt{<PoT>}, \texttt{</PoT>}, \texttt{<text2sql>}, \texttt{</text2sql>}, \texttt{<CoT>}, \texttt{</CoT>} \\
Execution Steps & \texttt{<N=$k$\_code>}, \texttt{</N=$k$\_code>}, \texttt{<N=$k$\_execution\_result>}, \texttt{</N=$k$\_execution\_result>} \newline (where $k \in \{0,1,2,3\}$) \\
For \textit{Sch} & \texttt{<NOTHING>} \\ \hline
\end{tabular}%
% }
\caption{List of structured special tokens added to DeBERTaV3-large tokenizer.}
\label{tab:special_tokens}
\end{table*}

These tokens structured all debugging results from \textit{CoTA}, \textit{PoTA}, \textit{t2SA}, \textit{PDA}, and \textit{SDA}, preparing them as inputs for fine-tuning DeBERTaV3-large.
The <NOTHING> token serves a special role. Any reasoning path containing this token is guaranteed to be incorrect, thus receiving a training confidence score of 0. The <NOTHING> token is used in two specific scenarios: first, reasoning paths deemed unnecessary by the scheduler are marked with <NOTHING>. For example, if the scheduler decides to execute text2SQL first and the CoT and text2SQL predictions match, PoT inference is skipped, and the corresponding position is forcibly filled with the <NOTHING> token. Consequently, the skipped PoT path is represented by <NOTHING>, and the CC effectively evaluates only CoT and text2SQL.

Additionally, some poorly performing LLMs occasionally generate irrelevant long token sequences when presented with excessively large tables or difficult questions. Such results hinder training and unnecessarily consume computational resources. Therefore, if the reasoning output exceeds 3000 characters, it is discarded and replaced by the <NOTHING> token, allowing the CC to efficiently assign lower confidence scores.

In the Figure~\ref{prompt:CC_nothing} and Figure~\ref{prompt:CC}, we present examples of input texts to the Confidence Checker that include the added special tokens.

% \newtcolorbox{mybox}[1][]{colback=white, colframe=black!75, fonttitle=\bfseries, title=#1}
\begin{figure*}[t!]
\centering
\begin{tcolorbox}[title=An Example Input Format for the Confidence Checker]
<Table\_row\_size>5</Table\_row\_size> \\
<Table\_column\_size>2</Table\_column\_size> \\
<Table\_size>10</Table\_size> \\
<Table> \\
| Day ~~~~~~~~~~~~|   Boxes of cookies | \\
|:---------------|---------------------:| \\
| Tuesday~~~~~~~|~~~~~~~~~~~~~~~~~~~~~~~ 25 | \\
| Wednesday~~|~~~~~~~~~~~~~~~~~~~~~~~ 27 | \\
| Thursday~~~~~|~~~~~~~~~~~~~~~~~~~~~~~ 23 | \\
| Friday~~~~~~~~~~|~~~~~~~~~~~~~~~~~~~~~~~~26 | \\
| Saturday~~~~~~|~~~~~~~~~~~~~~~~~~~~~~~ 23 | \\
</Table> \\
<Question>A Girl Scout troop recorded how many boxes of cookies they sold each day for a week. According to the table, what was the rate of change between Wednesday and Thursday?</Question> \\
<PoT> \\
<N=0\_code>\\
\# Calculate the rate of change between Wednesday and Thursday \\
rate\_of\_change = df.loc[1, `Boxes of cookies'] - df.loc[2, `Boxes of cookies'] \\
ans = rate\_of\_change</N=0\_code> \\
<N=0\_execution\_result>4</N=0\_execution\_result> \\
<N=1\_code> \\
\# Calculate the rate of change between Wednesday and Thursday \\
rate\_of\_change = df.loc[1, `Boxes of cookies'] - df.loc[2, `Boxes of cookies'] \\
ans = rate\_of\_change</N=1\_code> \\
<N=1\_execution\_result>4</N=1\_execution\_result> \\
</PoT> \\
<text2sql> \\
<N=0\_code>SELECT (b.Boxes\_of\_cookies - a.Boxes\_of\_cookies) AS answer \\
FROM dataframe a \\
JOIN dataframe b ON a.Day = `Wednesday' AND b.Day = `Thursday';</N=0\_code> \\
<N=0\_execution\_result>[[-4]]</N=0\_execution\_result> \\
</text2sql> \\
<CoT> \\
<solution>To find the rate of change in the number of boxes sold between Wednesday and Thursday, we need to calculate the difference in the number of boxes sold on these days. \\
Step 1: Identify the values from the table: \\
- Boxes sold on Wednesday = 27 \\
- Boxes sold on Thursday = 23 \\
Step 2: Calculate the rate of change by subtracting the number of boxes sold on Thursday from the number sold on Wednesday: \\
Rate of Change = Boxes sold on Wednesday - Boxes sold on Thursday = 27 - 23 = 4 \\
The rate of change is a decrease of 4 boxes.</solution> \\
<answer>-4</answer> \\
</CoT>
\end{tcolorbox}
\caption{An Example Input Format for the Confidence Checker}
\label{prompt:CC_nothing}
\end{figure*}

\begin{figure*}[t!]
\centering
\begin{tcolorbox}[title=An Example Input Format for the Confidence Checker with the <NOTHING> Token]
<Table\_row\_size>5</Table\_row\_size> \\
<Table\_column\_size>2</Table\_column\_size> \\
<Table\_size>10</Table\_size> \\
<Table> \\
| Day ~~~~~~~~~~~~|   Boxes of cookies | \\
|:---------------|---------------------:| \\
| Tuesday~~~~~~~|~~~~~~~~~~~~~~~~~~~~~~~ 25 | \\
| Wednesday~~|~~~~~~~~~~~~~~~~~~~~~~~ 27 | \\
| Thursday~~~~~|~~~~~~~~~~~~~~~~~~~~~~~ 23 | \\
| Friday~~~~~~~~~~|~~~~~~~~~~~~~~~~~~~~~~~~26 | \\
| Saturday~~~~~~|~~~~~~~~~~~~~~~~~~~~~~~ 23 | \\
</Table> \\
<Question>A Girl Scout troop recorded how many boxes of cookies they sold each day for a week. According to the table, what was the rate of change between Wednesday and Thursday?</Question> \\
<PoT> \\
<N=0\_code><NOTHING></N=0\_code> \\
<N=0\_execution\_result><NOTHING></N=0\_execution\_result> \\
<N=1\_code><NOTHING></N=1\_code> \\
<N=1\_execution\_result><NOTHING></N=1\_execution\_result> \\
</PoT> \\
<text2sql> \\
<N=0\_code>SELECT (b.Boxes\_of\_cookies - a.Boxes\_of\_cookies) AS answer \\
FROM dataframe a \\
JOIN dataframe b ON a.Day = `Wednesday' AND b.Day = `Thursday';</N=0\_code> \\
<N=0\_execution\_result>[[-4]]</N=0\_execution\_result> \\
</text2sql> \\
<CoT> \\
<solution>To find the rate of change in the number of boxes sold between Wednesday and Thursday, we need to calculate the difference in the number of boxes sold on these days. \\
Step 1: Identify the values from the table: \\
- Boxes sold on Wednesday = 27 \\
- Boxes sold on Thursday = 23 \\
Step 2: Calculate the rate of change by subtracting the number of boxes sold on Thursday from the number sold on Wednesday: \\
Rate of Change = Boxes sold on Wednesday - Boxes sold on Thursday = 27 - 23 = 4 \\
The rate of change is a decrease of 4 boxes.</solution> \\
<answer>-4</answer> \\
</CoT>
\end{tcolorbox}
\caption{An Example Input Format for the Confidence Checker with the <NOTHING> Token}
\label{prompt:CC}
\end{figure*}

\subsubsection{Segmented Training for Large Tables}

Training DeBERTaV3-large with excessively large tables posed challenges due to GPU memory constraints. Therefore, tables exceeding a maximum token limit were segmented for training purposes. Specifically, the table schema, considered critical information, was maintained entirely without omissions. Instead, the table was segmented row-wise, without dividing columns. Additionally, each epoch introduced a sliding window, shifting the rows downward sequentially, resulting in augmented datasets per epoch.

Moreover, prior studies have shown that table size is a critical factor influencing performance in table understanding. Accordingly, we also include the number of rows, columns, and the overall size of the table as input features to the Confidence Checker. A more detailed analysis of table size can be found in Appendix~\ref{appendix:sectionF}.

Training utilized two 80GB NVIDIA A100 GPUs with a batch size of 18 per device. The initial learning rate was set at 1e-6, employing a cosine learning rate scheduler and weight decay of 0.01 over 10 epochs.

\newpage
\vspace*{\fill}
\null

\newpage
\vspace*{\fill}
\null

\newpage
\vspace*{\fill}
\null

\newpage
\vspace*{\fill}
\null

\newpage

\section{Details of Benchmarks and Baselines, and Metrics}
\label{appendix:C}

\begin{table*}[h]
\centering
\scalebox{0.85}{
\begin{tabular}{l|c|c|c|c|c}
\toprule
\textbf{Dataset} & \textbf{\#Train} & \textbf{\#Val} & \textbf{\#Test} & \textbf{Main Task} & \textbf{Table Source} \\
\midrule
TableBench & -- & -- & 886 & Complex Table QA & Domain tables (finance, science, etc.) \\
penguins in a table & -- & -- & 144 & Basic Table Reasoning & Manually created toy table \\
\bottomrule
\end{tabular}
}
\caption{Summary of table QA datasets used in our experiments. Since we targeted only TableQA on a single table as our task, we removed cases that perform QA with multiple tables from Penguins in a Table, and refined 144 evaluation samples through a human evaluation check. Regarding TableBench, it contains heterogeneous target formats, including short entity or numeric answers, sentence-level descriptive outputs, and chart-generation tasks. Since our focus is entity-type TableQA, where the expected output is a concise entity or value, we evaluate the entity-answer subset of TableBench (693 out of 886 examples) and use EM, fuzzy matching, and token-level F1 as the main metrics. We exclude examples whose targets are not short entity-style answers, such as those in the Descriptive Analysis, Anomaly Detection, Causal Analysis, and Chart Generation categories.}
\label{tab:dataset-summary-benchmark}
\end{table*}

\subsection{Dataset}

This section provides a detailed description of the two benchmark datasets we used (see Table~\ref{tab:dataset-summary-benchmark}).

\noindent\textbf{\textbullet{} TableBench} is a comprehensive benchmark for complex table QA, spanning 18 subcategories across fact checking, numerical reasoning, data analysis, and visualization. Tables are sourced from diverse domains such as finance, sports, and science. The benchmark emphasizes real-world complexity and supports multiple reasoning paradigms including TCoT, SCoT, and PoT. In our experiments, since our primary focus is on entity-type TableQA where the expected output is a concise entity or value, we evaluated only the entity-answer subset of TableBench. Specifically, we excluded examples whose targets are not short entity-style answers, such as those in the Descriptive Analysis, Anomaly Detection, Causal Analysis, and Chart Generation categories. As a result, 693 out of the original 886 test samples are included in our evaluation, using EM, fuzzy matching, and token-level F1 as the main metrics.

\noindent\textbf{\textbullet{} Penguins in a Table} is a diagnostic dataset from BIG-bench designed to test basic table reasoning. It presents a single table of penguin species with attributes like height and weight, and asks simple factual or comparative questions. This isolates core table understanding without involving complex language or multi-step logic. Since we targeted only TableQA on a single table as our task, we removed cases that perform QA with multiple tables. After refining the samples through a human evaluation check, a final set of 144 evaluation samples was used.

\subsection{Baselines}

\noindent\textbf{\textbullet{} TabLaP}~\cite{TabLaP} is a multi-LLM system for table QA that delegates numerical reasoning to Python code execution. It prompts an LLM (NumSolver) to generate a Python script for answering numerical questions, while using a SOTA TableQA model for non-numerical ones. A separate LLM-based module (AnsSelector) selects the more reliable answer between the two branches, and a trustworthiness evaluator (TwEvaluator) estimates answer reliability to support regret-aware usage.

\noindent\textbf{\textbullet{} SynTQA}~\cite{zhang-etal-2024-syntqa} is an ensemble framework that combines Text-to-SQL and end-to-end table QA (E2E TQA) models by selecting the more reliable answer from both. It leverages the complementary strengths of each approach—Text-to-SQL excels at numerical reasoning and long tables, while E2E TQA is better at handling ambiguous questions and complex table content. A lightweight selector (either feature-based or LLM-based) chooses the final output, achieving improved accuracy and robustness over individual models.

\noindent\textbf{\textbullet{} MixSC}~\cite{Mix-SC} is a table QA framework that combines textual and symbolic reasoning via a mix self-consistency mechanism. It uses GPT-3.5 to perform both direct prompting and Python code execution (via a shell agent), then aggregates multiple outputs from each reasoning path to improve answer robustness. A normalization module (NORM) further enhances stability against structural perturbations like table transposition and row shuffling.

\subsection{Examples from the \textbf{Penguins in a Table}}

\begin{figure*}[t!]
    \centering
    \begin{subfigure}[t]{0.48\linewidth}
        \includegraphics[width=\linewidth]{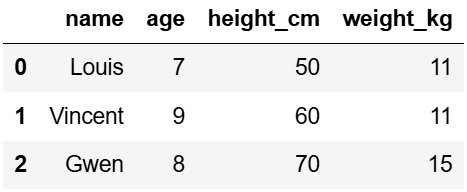}
        \caption{Example of \textbf{Penguins in a Table} (Index=10).\\
        Q: How many penguins are more than 8 years old?\\
        A: 1}
        \label{fig:penguinsexample1}
    \end{subfigure}
    \hfill
    \begin{subfigure}[t]{0.48\linewidth}
        \includegraphics[width=\linewidth]{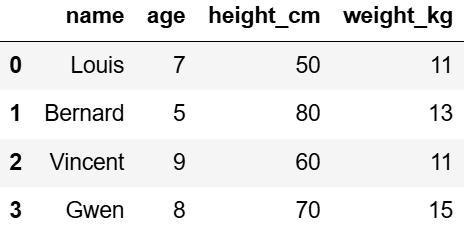}
        \caption{Example of \textbf{Penguins in a Table} (Index=140).\\
        Q: Which penguin is younger but taller than Gwen?\\
        A: Bernard}
        \label{fig:penguinsexample2}
    \end{subfigure}
    \caption{Two examples from the \textbf{Penguins in a Table} dataset.}
    \label{fig:penguins_combined}
\end{figure*}

% \begin{figure}[h]
%     \includegraphics[width=\linewidth]{images_appendix/penguins_14.png}
%     \caption{Example of \textbf{Penguins in a Table} dataset (Index=14). The question (Q) is \textbf{How many animals are listed in the table?}, the answer (A) is \textbf{4}}
% \label{fig:penguinsexample1}
% \end{figure}

% \begin{figure}[h]
%     \includegraphics[width=\linewidth]{images_appendix/penguins_140.png}
%     \caption{Example of \textbf{Penguins in a Table} dataset (Index=140). The question (Q) is \textbf{Which penguin is younger but taller than Gwen?}, the answer (A) is \textbf{Bernard}}
% \label{fig:penguinsexample2}
% \end{figure}

In the TableQA example shown in Figure~\ref{fig:penguinsexample1},  
the question (Q) is ``How many penguins are more than 8 years old?'',  
the answer (A) is \textbf{1}. 
In the second TableQA example shown in Figure~\ref{fig:penguinsexample2},  
the question (Q) is ``Which penguin is younger but taller than Gwen?'', the answer (A) is \textbf{Bernard}.

\subsection{Examples from the TableBench}
\label{tablebenchexamples}

\begin{figure*}[t!]
    \centering
    \includegraphics[width=\textwidth]{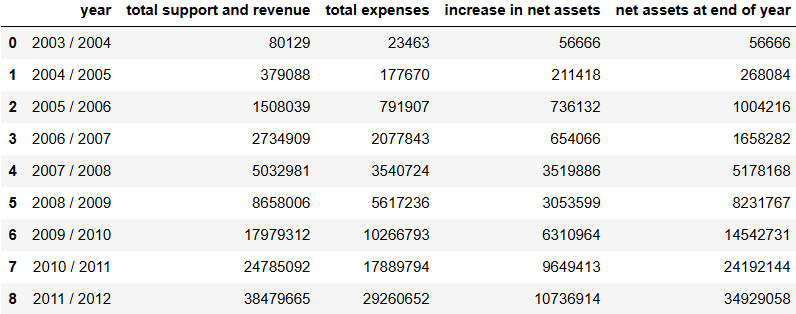}
    \caption{Example of \textbf{TableBench} dataset (Index=385). The question (Q) is \textbf{What is the total increase in net assets over the 3-year period from 2005/2006 to 2007/2008?}, the answer (A) is \textbf{4910084}}
\label{fig:tablebenchexample1}
\end{figure*}

\begin{figure*}[t!]
    \centering
    \includegraphics[width=\textwidth]{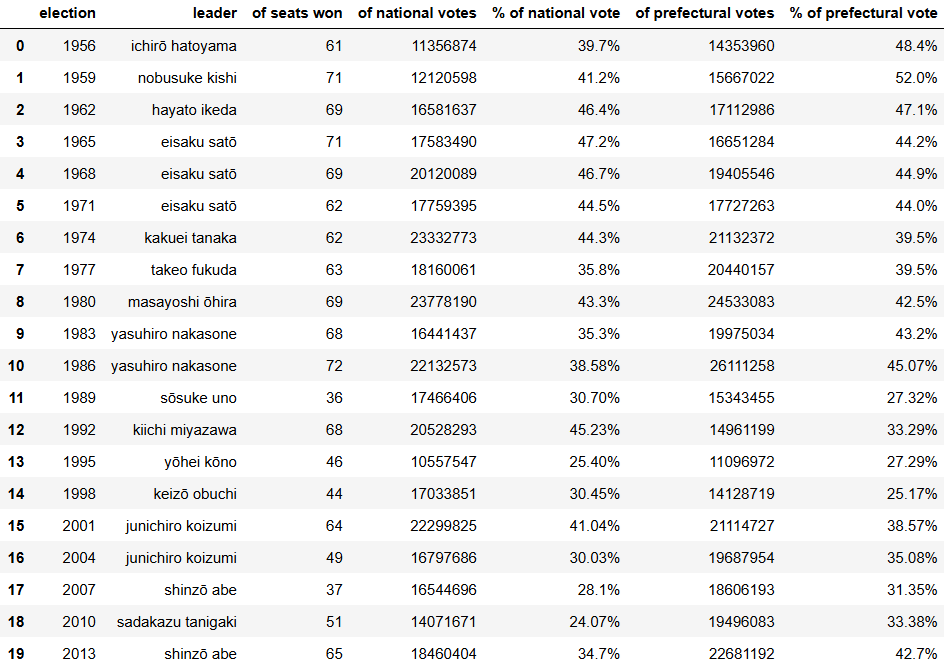}
    \caption{Example of \textbf{TableBench} dataset (Index=8). The question (Q) is \textbf{What is the average percentage of national votes won by all leaders in the table?}, the answer (A) is \textbf{37.64\%}}
\label{fig:tablebenchexample2}
\end{figure*}

In the TableQA example shown in Figure~\ref{fig:tablebenchexample1},  
the question (Q) is ``What is the total increase in net assets over the 3-year period from 2005/2006 to 2007/2008?'',  
the answer (A) is \textbf{4910084},  
and the question type is \texttt{Time-basedCalculation}. 
In the second TableQA example shown in Figure~\ref{fig:tablebenchexample2},  
the question (Q) is ``What is the average percentage of national votes won by all leaders in the table?'', the answer (A) is \textbf{37.64\%},  
and the question type is \texttt{aggregation}.

% \subsection{Worked example about MATA} 
% For the TableBench example in Figure~\ref{fig:tablebenchexample1}, MATA first runs \textit{CoTA} and uses \textit{Sch} to prioritize the code-based branch. The selected PoT/text2SQL branch generates executable code and, if needed, applies the corresponding debug agent. The resulting CoT, PoT, and text2SQL candidates are then scored by \textit{CC}; if the maximum confidence exceeds $\theta$, MATA selects the highest-scoring answer without \textit{JA}, otherwise \textit{JA} resolves the candidates using the table, question, and reasoning traces.

\subsection{Metrics}

% \textcolor{red}{

In many previous TableQA studies, Exact Match (EM) accuracy has been the most commonly used evaluation metric. However, EM alone is insufficient for assessing the quality of answers generated by LLMs.  For instance, if the ground truth is “John and Andy” but the model outputs “John , Andy” EM would consider it incorrect, even though the answer is semantically accurate. 

Therefore, unlike prior work, we employed two additional metrics alongside EM. First, we used fuzzy matching\footnote{https://pypi.org/project/fuzzywuzzy/}, a metric widely adopted in various studies \cite{king-flanigan-2024-unsupervised, cheng-etal-2024-dataflow, nekvinda-dusek-2021-shades} to measure textual similarity based on Levenshtein distance \cite{levenshtein1966binary}. Second, we adopted the SQuAD-style token-level F1 score \cite{rajpurkar-etal-2016-squad}, which evaluates token-level overlap between the prediction and the ground truth. By incorporating these two metrics, we complement the strictness of EM with more flexible and nuanced evaluations.

\newpage

\subsubsection{Fuzzy Matching Score for
           \texttt{``John , Andy''} vs.\ \texttt{``John and Andy''}}

Given two candidate answers\footnote{All strings are lower-cased and stripped of leading/trailing whitespace before comparison.}
\vspace{-0.5\baselineskip}
\begin{quote}
\begin{tabular}{@{}ll@{}}
$s_1 =$ & ``\texttt{john , andy}'' \quad\,(length $=11$)\\
$s_2 =$ & ``\texttt{john and andy}'' \quad\,(length $=13$)
\end{tabular}
\end{quote}

The similarity score is computed by \texttt{fuzz.ratio()}, which
defaults to \texttt{difflib.SequenceMatcher} when
\texttt{python‐Levenshtein} is \emph{not} installed.  The score is
\[
\text{SM}(s_1,s_2)=\frac{2M}{|s_1|+|s_2|}\times 100,
\]
where $M$ is the total length of all matching blocks returned by
\texttt{SequenceMatcher}.  
For the strings above, the matching blocks are
``\texttt{john }'' (length~$5$) and ``\texttt{ andy}'' (length~$5$) with whitespace characters counted in the length.
Thus, $M = 5 + 5 = 10$, and the final similarity score is
\[
\text{SM}(s_1, s_2) = \frac{2 \times 10}{11 + 13} \times 100 = 83.33\%.
\]
\texttt{fuzz.ratio()} rounds this to~83; dividing by~100 in our
post-processing step yields the final value $0.83$.

%--------------------------------------------------------------------
% SQuAD-style token-level F1 예시
%--------------------------------------------------------------------
\subsubsection{Token-level F1 Score for
           \texttt{``John , Andy''} vs.\ \texttt{``John and Andy''}}

Following the SQuAD\,v1.1 definition, each answer is first
\emph{lower-cased} and \emph{split on whitespace} to obtain tokens:
\begin{align*}
\text{prediction tokens}  &:= \{ \text{``john''},\; \text{``,''},\;
                                \text{``andy''} \}, \\[2pt]
\text{ground-truth tokens} &:= \{ \text{``john''},\; \text{``and''},\;
                                \text{``andy''} \}.
\end{align*}

\paragraph{Overlap.}
The set intersection contains \emph{two} common tokens,
``\texttt{john}'' and ``\texttt{andy}'', so
\[
n_{\text{same}} = 2.
\]

\paragraph{Precision and recall.}
\[
\text{precision}
= \frac{n_{\text{same}}}{|\text{prediction tokens}|}
= \frac{2}{3},
\]

\[
\text{recall}
= \frac{n_{\text{same}}}{|\text{ground-truth tokens}|}
= \frac{2}{3}.
\]

\paragraph{F1 score.}
\begin{align*}
F_1
&= \frac{2 \times \text{precision} \times \text{recall}}
         {\text{precision} + \text{recall}} \\
&= \frac{2 \times \tfrac23 \times \tfrac23}{\tfrac23 + \tfrac23} = \frac23 \approx 0.67.
\end{align*}
% \[
% F_1
% = \frac{\,2 \times \text{precision} \times \text{recall}\,}
%        {\,\text{precision} + \text{recall}\,}
% = \frac{2 \times \tfrac23 \times \tfrac23}{\tfrac23 + \tfrac23}
% = \frac23
% \approx 0.67.
% \]

\noindent
Thus, the SQuAD-style token-level F1 score for the two example strings is
\[
\boxed{F_1 = 0.67}.
\]

\section{Hyper-Parameter Setting}
\label{appendix:E}

This section reports the experimental results that guided the selection of MATA’s hyperparameters, $N$ and $\theta$. Based on these results, MATA sets $N = 3$ and $\theta = 0.1$.

\subsection{Debugging Iteration Stop Condition}

Unlike CoT, PoT and text2SQL often suffer from execution failures, particularly when encountering atypical table structures or inconsistent value types. To address this issue, we apply a lightweight self-refinement strategy \cite{self-refine} for PoT and text2SQL (rather than a full-fledged LLM-based code debugging\cite{self-debugging, Debug-like-a-Human}). Each generated code snippet, along with its execution result, is re-injected into the LLM for iterative correction, up to a maximum of $N$ rounds. Early stopping is applied when refinements show minimal change or execution outcomes remain unchanged, thereby reducing unnecessary computation.

For Python code generated via PoT, refinement termination is determined by computing four similarity metrics including Levenshtein distance~\cite{levenshtein1966binary}, difflib similarity~\cite{python-difflib, ratcliff1988pattern}, Abstract Syntax Tree (AST) similarity~\cite{AbstractSyntaxTree}, and Opcode-based similarity~\cite{Opcode}. Refinement is considered complete when the average of these metrics exceeds 0.9, indicating convergence.

The self-refinement process for text2SQL follows a distinct stopping criterion. Unlike Python code, SQL queries tend to undergo minimal changes during refinement and often resist correction when semantic or structural errors are present. Accordingly, we stop the refinement loop for text2SQL either when the generated query remains unchanged across iterations or when previously failing queries execute successfully. Further details are described in Algorithm~\ref{algo:mata}.

Across both PoT and text2SQL, up to $N$ self-refinement rounds may be applied depending on the stability of intermediate outputs. CoT, on the other hand, is excluded from refinement. Empirically, we observe that CoT solutions are stable across repeated generations and rarely benefit from additional passes. To avoid unnecessary LLM calls, we retain a single-step inference for CoT (See Appendix~\ref{sec:cotN}).

At the end of this process, MATA obtains one text-based answer from CoT, along with up to (1+$N$) candidate outputs each from PoT and text2SQL, including code executions. These are then aggregated and passed to the Confidence Checker for final decision-making. We set $N=3$, by default, in all experiments. 

\begin{figure*}[b!]
    \centering
    \includegraphics[width=0.7\textwidth]{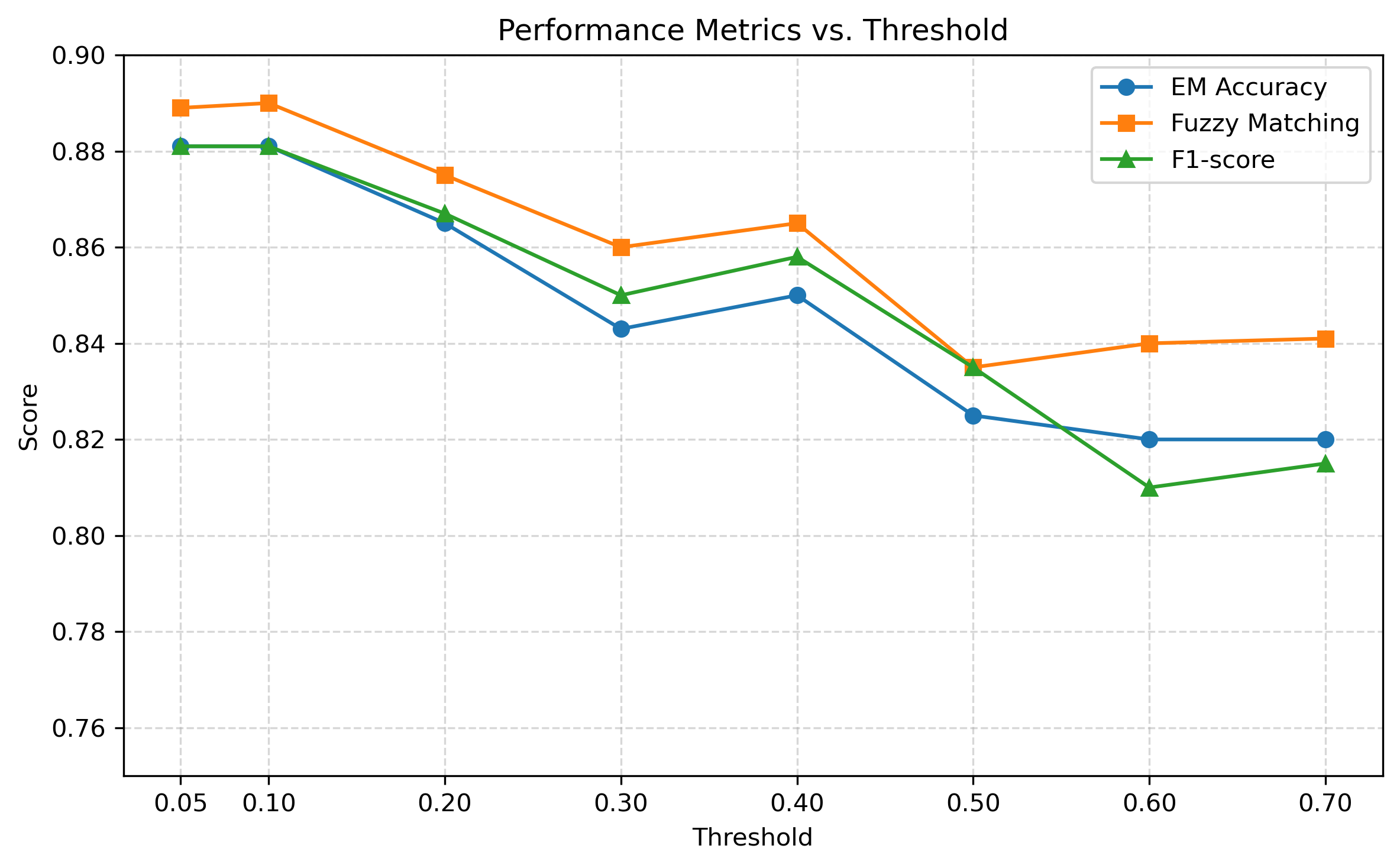}
    \caption{Performance scores (EM, fuzzy, F1) averaged over 10 models on the \textbf{Penguins in a Table}. As the confidence threshold $\theta$ increases, performance gradually decreases.}
\label{fig:penguins_threshold}
\end{figure*}

\begin{figure*}[b!]
    \centering
    \includegraphics[width=0.7\textwidth]{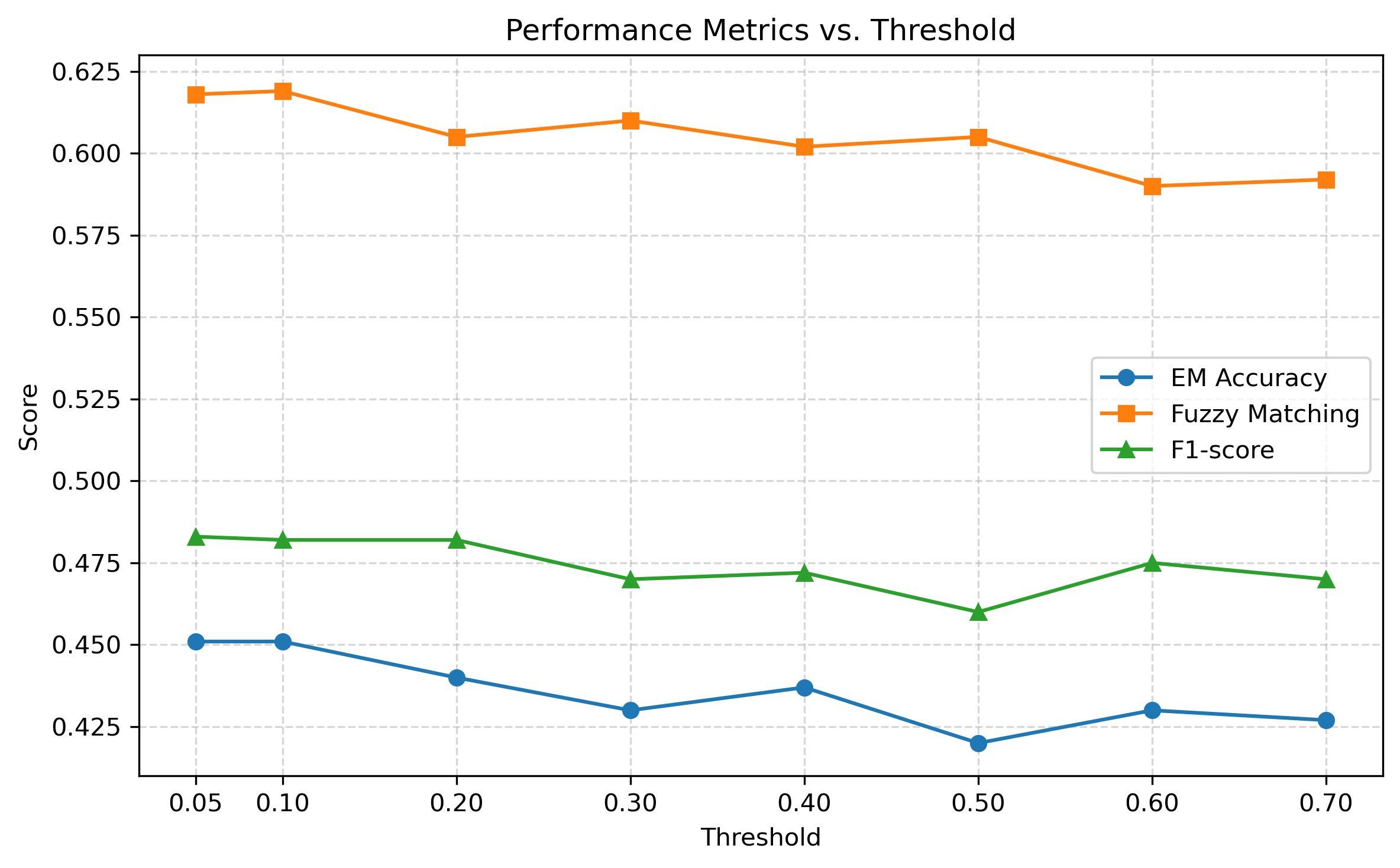}
    \caption{Performance scores (EM, fuzzy, F1) averaged over 10 models on the \textbf{TableBench}. Higher values of $\theta$ result in more frequent fallback to the Judge Agent, leading to degraded performance.}
\label{fig:tablebench_threshold}
\end{figure*}

\subsection{The Maximum Number Of Debugging Iteration ($N$)}
\label{sec:iter}

Tables~\ref{tab:PoTDebugStopPenguin}, \ref{tab:PoTDebugStopTablebench}, \ref{tab:text2SQLDebugStopPenguin} and \ref{tab:text2SQLDebugStopTablebench} present the iteration at which the debug loop terminates early when the maximum number of iterations is set to $N = 7$. The results show that for nearly all models, debugging for both PoT and text2SQL typically concludes within just one or two iterations. Additional iterations beyond that point generally yield little benefit. An exception is Mistral-7B, which we attribute to its limited debugging capability via self-refinement \cite{self-refine}. Based on these findings, we set the default maximum number of iterations to $N = 3$ in order to minimize unnecessary LLM agent calls while still enabling high-quality PoT and text2SQL reasoning through effective debugging.

% \newpage

% \newpage

\subsection{CoT Self-Refinement Behavior and Early Stopping Analysis}
\label{sec:cotN}

To assess the inefficiency of the debug iteration loop for CoT, we implemented a CoT Debug Agent (\textit{CDA}) using the same prompt structure as the Python Debug Agent (\textit{PDA}) and SQL Debug Agent (\textit{SDA}) (see Appendix~\ref{appendix:sectionG} for prompt details). Like the others, \textit{CDA} employs self-refinement, and we set $N = 7$ as in Appendix~\ref{sec:iter} to observe how the CoT solution and answer are revised during the process. The experimental results are shown in Table~\ref{tab:CoTDebugStop_penguin} and Table~\ref{tab:CoTDebugStop_tablebench}. The results show that in most cases, the iteration stops at $N = 1$, indicating that the initial CoT reasoning generated by \textit{CoTA} is rarely altered by \textit{CDA}. Therefore, to maximize the efficiency of MATA, we chose not to introduce \textit{CDA} alongside \textit{CoTA}.

% \newpage

\subsection{Effect of Confidence Threshold ($\theta$) on MATA's Performance} 
\label{appendix:sweep_theta}

We input the solutions, answers, code, and code execution results generated by CoT, PoT, and text2SQL reasoning into the Confidence Checker (\textit{CC}) to obtain confidence scores for each reasoning path. If the confidence score is sufficiently high, the system selects the reasoning path with the highest score as the final answer without invoking the additional Judge Agent (\textit{JA}). This allows us to obtain high-quality answers while avoiding extra LLM agent calls. The threshold $\theta$, which determines whether a confidence score is deemed sufficiently high, is a tunable hyperparameter. We investigated the effect of varying this hyperparameter.

Figure~\ref{fig:penguins_threshold} and Figure~\ref{fig:tablebench_threshold} show that increasing $\theta$ generally leads to a decrease in overall performance. This aligns with the analysis in the main Ablation Study: as $\theta$ increases, the final answer is more often determined by JA rather than \textit{CC}. Given that removing \textit{CC} significantly reduced performance in the ablation study, we infer that \textit{CC} is more effective than JA in selecting the correct answer. Thus, increasing $\theta$ leads to more frequent (and less effective) reliance on JA, resulting in lower performance. When $\theta$ is set below 0.1, we observed no significant change in performance, suggesting that confidence scores below 0.1 are not meaningful. Therefore, we conclude that setting $\theta = 0.1$ is an appropriate choice.

% \newpage

\subsection{Confidence Threshold Analysis}
\label{sec:confidence_threshold}

We further analyze whether the Confidence Checker (\textit{CC}) provides a reliable signal for cost-aware early exiting. For each query, \textit{CC} assigns three confidence scores to the candidate answers produced by CoT, PoT, and text2SQL, denoted as C, P, and TS, respectively. We define $s=\max(\text{C},\text{P},\text{TS})$ as the confidence of the answer that would be selected by \textit{CC}. If $s>\theta$, MATA accepts the highest-scoring candidate and skips the \textit{JA}; otherwise, the query is treated as uncertain and passed to \textit{JA}.

The goal of \textit{CC} is not to produce perfectly calibrated probabilities, but to serve as a lightweight gating and ranking module that decides when \textit{JA} can be safely skipped. We therefore evaluate whether $s$ separates cases where the \textit{CC}-selected answer is likely to be correct from cases where additional verification is needed. Following the threshold sweep in Appendix~\ref{appendix:sweep_theta}, we use $\theta=0.1$: thresholds below 0.1 showed little effect, while larger thresholds increased fallback to \textit{JA} and generally degraded final performance.

\begin{table}[h!]
\centering
\small
\begin{tabular}{p{2.5cm}|p{1.2cm}p{1.2cm}p{1.0cm}}
\toprule
Metric & llama3.2-3b & qwen2.5-32b & GPT-4o \\
\midrule
ROC-AUC $\uparrow$              & 0.875 & 0.886 & 0.869 \\
ECE$_{10}$ $\downarrow$         & 0.163 & 0.154 & 0.162 \\
\textit{JA}-skip@$\theta=0.1$ $\uparrow$ & 0.577 & 0.698 & 0.730 \\
EM($s>\theta$) $\uparrow$       & 0.545 & 0.698 & 0.700 \\
EM($s\leq\theta$) $\uparrow$    & 0.092 & 0.177 & 0.267 \\
\bottomrule
\end{tabular}
\caption{Confidence-score analysis on \textbf{TableBench} ($n=693$). \textit{CC} outputs confidence scores C, P, and TS for CoT, PoT, and text2SQL, and we define $s=\max(\text{C},\text{P},\text{TS})$. ROC-AUC evaluates whether $s$ ranks correct \textit{CC}-selected answers above incorrect ones. ECE$_{10}$ is the 10-bin expected calibration error. \textit{JA}-skip@$\theta=0.1$ is the fraction of queries where $s>\theta$, so \textit{JA} is skipped. EM($s>\theta$) and EM($s\leq\theta$) report the EM of the \textit{CC}-selected answer in the high- and low-confidence partitions.}
\label{tab:cc_calibration}
\end{table}

% \begin{table}[t]
% \centering
% \small
% \setlength{\tabcolsep}{4pt}
% \begin{tabular}{lccccc}
% \toprule
% Model & \begin{tabular}{@{}c@{}}ROC\\-AUC\end{tabular} & ECE$_{10}$ & \begin{tabular}{@{}c@{}}JA-skip\\@$\theta=0.1$\end{tabular} & \begin{tabular}{@{}c@{}}EM\\($s>\theta$)\end{tabular} & \begin{tabular}{@{}c@{}}EM\\($s\leq\theta$)\end{tabular} \\
% \midrule
% Llama3.2-3B & 0.875 & 0.163 & 0.577 & 0.545 & 0.092 \\
% Qwen2.5-32B & 0.886 & 0.154 & 0.698 & 0.698 & 0.177 \\
% GPT-4o      & 0.869 & 0.162 & 0.730 & 0.700 & 0.267 \\
% \bottomrule
% \end{tabular}
% \caption{Confidence-score quality on TableBench ($n=693$). We evaluate $s=\max(C,P,TS)$ as a predictor of whether the CC-selected answer is correct. The threshold $\theta=0.1$ separates confident and uncertain cases while enabling substantial Judge Agent skipping.}
% \label{tab:cc_calibration}
% \end{table}

Table~\ref{tab:cc_calibration} reports this analysis on \textbf{TableBench}. ROC-AUC measures how well $s$ ranks correct \textit{CC}-selected answers above incorrect ones. ECE$_{10}$ is the expected calibration error computed with 10 bins, where lower values indicate better calibration. \textit{JA}-skip@$\theta=0.1$ denotes the fraction of examples for which $s>\theta$, meaning that MATA accepts the \textit{CC}-selected answer without invoking \textit{JA}. Finally, EM($s>\theta$) and EM($s\leq\theta$) report the exact-match accuracy of the \textit{CC}-selected answer in the high- and low-confidence partitions, respectively.

Across representative small, large open-source, and closed-source backbones, \textit{CC} achieves strong ROC-AUC scores of 0.869--0.886. More importantly, the high-confidence group consistently obtains much higher EM than the low-confidence group: for example, 0.698 vs. 0.177 with qwen2.5-32B and 0.700 vs. 0.267 with GPT-4o. At the same time, MATA skips \textit{JA} for 57.7--73.0\% of \textbf{TableBench} queries. These results support the use of $\theta=0.1$ as a practical early-exit threshold: it identifies many cases where \textit{CC} alone is reliable, while reserving \textit{JA} for genuinely uncertain cases.

\newpage

\section{Additional Experiments Results}
\label{appendix:F}

\subsection{Detailed Results of the Ablation Study}

Table~\ref{tab:ablipenguins} and Table~\ref{tab:ablitablebench} provide the exact numerical data corresponding to Figure~\ref{fig:ablation} in the main text.

\subsection{Reduction in LLM Agent Calls via Confidence Checker (\textit{CC}) and Scheduler (\textit{Sch})}

This section reports the experimental results on the reduction in LLM agent calls achieved by the \textit{CC} and \textit{Sch} modules, as discussed in the Ablation Study section of the main text. See Tables~\ref{tab:schCC_penguin} and \ref{tab:schCC_tablebench} for detailed results.
On the \textbf{Penguins in a Table} dataset, applying the \textit{CC} results in approximately a 95.8\% reduction in LLM agent calls across all 10 models, while using \textit{Sch} yields a reduction of about 14.59\%. On the \textbf{TableBench} dataset, the CC reduces LLM agent calls by approximately 60.6\%, and Sch by about 7.62\%, again averaged across all 10 models. Additionally, both \textit{CC} and \textit{Sch} were found to be more effective with large LLMs than with smaller ones. This suggests that these modules become even more beneficial in settings where large LLMs—incurring higher inference costs—are used as backbones.

\subsection{Additional TableBench Results with ROUGE-L and Protocol Clarification}
\label{app:rouge_l}

\begin{table}[h!]
\centering
\small
\begin{tabular}{p{2.4cm}|p{0.8cm}p{0.95cm}p{0.75cm}p{0.7cm}}
\toprule
Model & TabLaP & SynTQA & MixSC & MATA \\
\midrule
llama3.2-3b & 0.207 & 0.138 & 0.218 & \textbf{0.417} \\
mistral-7b & 0.219 & 0.281 & 0.234 & \textbf{0.353} \\
phi4-mini-3.8b & 0.226 & 0.288 & 0.284 & \textbf{0.334} \\
qwen2.5-3b & 0.244 & 0.263 & 0.241 & \textbf{0.346} \\
qwen2.5-7b & 0.138 & 0.342 & 0.234 & \textbf{0.428} \\
mistral-small-24b & 0.363 & 0.447 & 0.421 & \textbf{0.637} \\
cogito-32b & 0.521 & 0.484 & 0.514 & \textbf{0.633} \\
qwen2.5-32b & 0.380 & 0.446 & 0.394 & \textbf{0.634} \\
GPT-4o & 0.630 & 0.514 & 0.580 & \textbf{0.654} \\
Claude-3.7-Sonnet & 0.666 & 0.545 & \textbf{0.690} & 0.684 \\
\midrule
Average & 0.359 & 0.375 & 0.381 & \textbf{0.512} \\
\bottomrule
\end{tabular}
\caption{ROUGE-L results on the TableBench entity-answer subset used for the main evaluation. Higher is better.}
\label{tab:appendix_rouge_l}
\end{table}

TableBench contains heterogeneous output formats, including short entity/numeric answers, sentence-level descriptive outputs, and chart-generation tasks. Since our target task is entity-type TableQA, the main paper evaluates on the entity-answer subset (693/886 examples), where discrete-answer comparison is meaningful. The excluded subsets correspond to Descriptive Analysis, Anomaly Detection, Causal Analysis, and Chart Generation. To improve comparability with prior work, we additionally report ROUGE-L on the same subset in Table~\ref{tab:appendix_rouge_l}.

\subsection{Additional In-Distribution Evaluation on WikiTQ, TabMWP, and TabFact}

While the main paper emphasizes out-of-distribution evaluation on \textbf{TableBench} and \textbf{Penguins in a Table}, we additionally report results on WikiTQ, TabMWP, and TabFact to verify that the gains of MATA are not limited to distribution-shift settings. Table~\ref{tab:appendix_id_llama} reports full-test Exact Match results with Llama3.2-3B. In this setting, MATA remains competitive and generally strong on all three benchmark families.

\begin{table}[h!]
\centering
\resizebox{\columnwidth}{!}{
\begin{tabular}{p{1.9cm}|p{1.3cm}|p{1.4cm}|p{1.3cm}}
\toprule
llama3.2-3b & WikiTQ (4,344) & TabMWP (7,686) & TabFact (12,779) \\
\midrule
TabLaP & 0.220 & 0.171 & 0.599 \\
SynTQA & 0.144 & 0.228 & 0.451 \\
MixSC & 0.232 & 0.216 & 0.612 \\
\textbf{MATA} & \textbf{0.535} & \textbf{0.713} & \textbf{0.688} \\
\bottomrule
\end{tabular}}
\caption{Exact Match results on the full test splits of WikiTQ, TabMWP, and TabFact using llama3.2-3b.}
\label{tab:appendix_id_llama}
\end{table}

\begin{table*}[h]
\centering
\scalebox{0.7}{
% \small
\begin{tabular}{l||c|c|c||c|c|c||c|c|c||c|c|c||c|c|c} 
\toprule
& \multicolumn{3}{c|}{\textit{w/o sch}}& \multicolumn{3}{c|}{\textit{w/o CC}} & \multicolumn{3}{c|}{\textit{w/o JA}} & \multicolumn{3}{c|}{\textit{w/o FM}}& \multicolumn{3}{c}{\textbf{MATA}}\\

\midrule
Models  &
EM  &
fuzzy  &
F1  &
EM  &
fuzzy  &
F1  &
EM  &
fuzzy &
F1  &
EM &
fuzzy  &
F1  &
EM  &
fuzzy  &
F1  \\

\midrule
llama3.2-3b &
\textbf{0.736} &
\textbf{0.766} &
\textbf{0.736} &
0.549 &
0.591 &
0.557 &
\underline{0.715} &
\underline{0.737} &
\underline{0.715} &
\textbf{0.736} &
\textbf{0.766} &
\textbf{0.736} &
\textbf{0.736} &
\textbf{0.766} &
\textbf{0.736} \\

mistral-7b &
\textbf{0.875} &
\textbf{0.893} &
\textbf{0.875} &
0.625 &
0.667 &
0.643 &
\underline{0.861} &
0.879 &
\underline{0.861} &
\underline{0.861} &
\underline{0.880} &
\underline{0.861} &
\underline{0.861} &
\underline{0.880} &
\underline{0.861} \\

phi4-mini-3.8b &
\underline{0.826} &
\underline{0.852} &
\underline{0.826} &
0.590 &
0.648 &
0.590 &
\textbf{0.833} &
\textbf{0.863} &
\textbf{0.833} &
0.819 &
0.847 &
0.819 &
0.819 &
0.847 &
0.819 \\

% qwen2.5-coder-7b &
% 0.917 &
% 0.925 &
% 0.917 &
% 0.694 &
% 0.727 &
% 0.694 &
% \textbf{0.931} &
% \textbf{0.939} &
% \textbf{0.931} &
% \underline{0.924} &
% \underline{0.932} &
% \underline{0.924} &
% \underline{0.924} &
% \underline{0.932} &
% \underline{0.924} \\

qwen2.5-3b &
\textbf{0.882} &
\textbf{0.896} &
\textbf{0.882} &
0.688 &
0.715 &
0.688 &
0.854 &
0.872 &
0.854 &
\underline{0.868} &
\underline{0.883} &
\underline{0.868} &
\underline{0.868} &
\underline{0.883} &
\underline{0.868} \\

qwen2.5-7b  &
\underline{0.931} &
\underline{0.939} &
\underline{0.931} &
0.854 &
0.862 &
0.854 &
\textbf{0.951} &
\textbf{0.955} &
\textbf{0.951} &
\textbf{0.951} &
\textbf{0.955} &
\textbf{0.951} &
\textbf{0.951} &
\textbf{0.955} &
\textbf{0.951} \\

% qwen2.5-14b &
% \textbf{0.917} &
% \textbf{0.917} &
% \textbf{0.917} &
% 0.840 &
% 0.846 &
% 0.840 &
% \underline{0.910} &
% \underline{0.910} &
% \underline{0.910} &
% \underline{0.910} &
% \underline{0.910} &
% \underline{0.910} &
% \underline{0.910} &
% \underline{0.910} &
% \underline{0.910} \\

\midrule
mistral-small-24b &
\textbf{0.917} &
\textbf{0.917} &
\textbf{0.917} &
0.875 &
0.875 &
0.875 &
\underline{0.896} &
\underline{0.896} &
\underline{0.896} &
\underline{0.896} &
\underline{0.896} &
\underline{0.896} &
\underline{0.896} &
\underline{0.896} &
\underline{0.896} \\

cogito-32b &
\textbf{0.903} &
\textbf{0.903} &
\textbf{0.903} &
\underline{0.875} &
\underline{0.875} &
\underline{0.875} &
\textbf{0.903} &
\textbf{0.903} &
\textbf{0.903} &
\textbf{0.903} &
\textbf{0.903} &
\textbf{0.903} &
\textbf{0.903} &
\textbf{0.903} &
\textbf{0.903} \\

qwen2.5-32b &
\textbf{0.931} &
\textbf{0.931} &
\textbf{0.931} &
0.882 &
0.882 &
0.882 &
\underline{0.917} &
\underline{0.917} &
\underline{0.917} &
\underline{0.917} &
\underline{0.917} &
\underline{0.917} &
\underline{0.917} &
\underline{0.917} &
\underline{0.917} \\

GPT-4o &
\textbf{0.944} &
\textbf{0.944} &
\textbf{0.944} &
0.882 &
0.886 &
0.887 &
\underline{0.903} &
\underline{0.903} &
\underline{0.903} &
\underline{0.903} &
\underline{0.903} &
\underline{0.903} &
\underline{0.903} &
\underline{0.903} &
\underline{0.903} \\

Claude-3.7-Sonnet &
\textbf{0.958} &
\textbf{0.958} &
\textbf{0.958} &
0.903 &
0.907 &
0.907 &
\underline{0.951} &
\underline{0.951} &
\underline{0.951} &
\underline{0.951} &
\underline{0.951} &
\underline{0.951} &
\underline{0.951} &
\underline{0.951} &
\underline{0.951} \\

\midrule
\textit{Average} &
\textbf{0.890} &
\textbf{0.900} &
\textbf{0.890} &
0.774 &
0.793 &
0.777 &
0.876 &
0.886 &
0.877 &
\underline{0.881} &
\underline{0.890} &
\underline{0.881} &
\underline{0.881} &
\underline{0.890} &
\underline{0.881} \\

\bottomrule
\end{tabular}
}
\caption{Ablation study results on the \textbf{Penguins in a Table} benchmark. We report Exact Match(EM) accuracy, fuzzy matching, and F1 scores for each model. Bold indicates the best performance; underlined scores are the second best.}
\label{tab:ablipenguins}
\end{table*}
\begin{table*}[h]
\centering
\scalebox{0.7}{
% \small
\begin{tabular}{l||c|c|c||c|c|c||c|c|c||c|c|c||c|c|c} 
\toprule
& \multicolumn{3}{c|}{\textit{w/o sch}}& \multicolumn{3}{c|}{\textit{w/o CC}} & \multicolumn{3}{c|}{\textit{w/o JA}} & \multicolumn{3}{c|}{\textit{w/o FM}}& \multicolumn{3}{c}{\textbf{MATA}}\\

\midrule
Models  &
EM  &
fuzzy  &
F1  &
EM  &
fuzzy  &
F1  &
EM  &
fuzzy &
F1  &
EM &
fuzzy  &
F1  &
EM  &
fuzzy  &
F1  \\

\midrule
llama3.2-3b &
\textbf{0.359} &
\textbf{0.571} &
\textbf{0.388} &
0.235 &
0.471 &
0.265 &
\underline{0.354} &
0.554 &
0.374 &
0.352 &
0.557 &
0.379 &
\underline{0.354} &
\underline{0.563} &
\underline{0.381} \\

mistral-7b &
\underline{0.293} &
\underline{0.471} &
\textbf{0.321} &
0.224 &
0.421 &
0.253 &
0.289 &
0.455 &
\underline{0.314} &
0.281 &
0.454 &
0.306 &
\textbf{0.294} &
\textbf{0.473} &
\textbf{0.321} \\

phi4-mini-3.8b &
\underline{0.271} &
\underline{0.458} &
\textbf{0.295} &
0.201 &
0.398 &
0.221 &
0.268 &
\textbf{0.460} &
0.294 &
0.270 &
0.440 &
0.288 &
\textbf{0.273} &
0.457 &
\textbf{0.295} \\

% qwen2.5-coder-7b &
% \textbf{0.446} &
% \textbf{0.625} &
% \textbf{0.481} &
% 0.325 &
% 0.545 &
% 0.361 &
% 0.433 &
% 0.608 &
% 0.463 &
% 0.442 &
% \underline{0.614} &
% 0.472 &
% \underline{0.444} &
% \textbf{0.625} &
% \underline{0.479} \\

qwen2.5-3b &
\textbf{0.293} &
\underline{0.469} &
\textbf{0.317} &
0.208 &
0.395 &
0.237 &
0.284 &
0.465 &
0.309 &
\underline{0.291} &
0.458 &
\underline{0.312} &
\underline{0.291} &
\textbf{0.471} &
\textbf{0.317} \\

qwen2.5-7b  &
\textbf{0.354} &
\textbf{0.557} &
\textbf{0.394} &
0.343 &
\underline{0.553} &
0.386 &
\underline{0.349} &
\underline{0.553} &
\underline{0.387} &
\underline{0.349} &
0.530 &
0.379 &
\textbf{0.354} &
\textbf{0.557} &
\underline{0.393} \\

% qwen2.5-14b &
% \textbf{0.511} &
% \textbf{0.678} &
% \textbf{0.545} &
% 0.463 &
% 0.655 &
% 0.499 &
% 0.502 &
% 0.670 &
% 0.534 &
% 0.505 &
% 0.664 &
% 0.532 &
% \underline{0.506} &
% \underline{0.674} &
% \underline{0.539} \\

\midrule
mistral-small-24b &
\textbf{0.573} &
\underline{0.722} &
\underline{0.604} &
\underline{0.540} &
0.702 &
0.576 &
0.532 &
0.697 &
0.568 &
\textbf{0.573} &
0.715 &
0.601 &
\textbf{0.573} &
\textbf{0.723} &
\textbf{0.605} \\

cogito-32b &
\textbf{0.577} &
\textbf{0.723} &
\textbf{0.609} &
0.514 &
0.691 &
0.552 &
0.564 &
\underline{0.713} &
0.596 &
\underline{0.571} &
0.711 &
\underline{0.600} &
\textbf{0.577} &
\textbf{0.723} &
\textbf{0.609} \\

qwen2.5-32b &
0.573 &
\underline{0.718} &
\underline{0.603} &
0.554 &
0.710 &
0.585 &
0.544 &
0.699 &
0.577 &
\underline{0.576} &
0.712 &
0.601 &
\textbf{0.577} &
\textbf{0.721} &
\textbf{0.607} \\

GPT-4o &
0.587 &
\underline{0.735} &
0.622 &
0.586 &
0.733 &
0.624 &
0.584 &
0.731 &
0.616 &
\underline{0.593} &
\underline{0.735} &
\underline{0.625} &
\textbf{0.595} &
\textbf{0.740} &
\textbf{0.629} \\

Claude-3.7-Sonnet &
\textbf{0.620} &
\textbf{0.767} &
\textbf{0.666} &
0.587 &
0.752 &
0.651 &
0.612 &
0.763 &
0.656 &
\underline{0.618} &
0.750 &
0.657 &
\textbf{0.620} &
\underline{0.764} &
\underline{0.664} \\

\midrule
\textit{Average} &
\underline{0.450} &
\textbf{0.619} &
\textbf{0.482} &
0.399 &
0.583 &
0.435 &
0.438 &
\underline{0.609} &
0.469 &
0.447 &
0.606 &
\underline{0.475} &
\textbf{0.451} &
\textbf{0.619} &
\textbf{0.482} \\

\bottomrule
\end{tabular}
}
\caption{Ablation study results on the \textbf{TableBench} benchmark. Bold and underline follow Table~\ref{tab:ablipenguins}.}
\label{tab:ablitablebench}
\end{table*}

\begin{table*}[h]
\centering
% \small
\begin{tabular}{l|c|c|c|c|c|c} 
\toprule

& \multicolumn{3}{c|}{\textit{Penguins in a Table}} & \multicolumn{3}{c} {\textit{TableBench}} \\

\midrule

\textbf{Method} &
\textbf{EM}  &
\textbf{fuzzy}  &
\textbf{F1}  &
\textbf{EM}  &
\textbf{fuzzy}  &
\textbf{F1}   \\

\midrule

ReAcTable &
0.653 &
0.705 &
0.738 &
\underline{0.338} &
\underline{0.471} &
\underline{0.402} \\

Chameleon &
\underline{0.882} &
\underline{0.902} &
\textbf{0.905} &
0.267 &
0.328 &
0.273 \\

\midrule
MATA&
\textbf{0.903} &
\textbf{0.903} &
\underline{0.903} &
\textbf{0.595} &
\textbf{0.740} &
\textbf{0.629} \\

\bottomrule
\end{tabular}
\caption{Evaluation results on the \textbf{Penguins in a Table} (left) and \textbf{TableBench} (right) datasets.
All methods use GPT-4o as the backbone. Bold indicates the best performance; underlined scores are the second best.}
\label{tab:gpt4o}
\end{table*}
\begin{table*}[t!]
\centering
\small
\begin{tabular}{llcccccc}
\toprule
Benchmark & Model & \multicolumn{3}{c}{AutoPrep} & \multicolumn{3}{c}{MATA} \\
\cmidrule(lr){3-5}\cmidrule(lr){6-8}
 &  & EM & fuzzy & F1 & EM & fuzzy & F1 \\
\midrule
Penguins in a Table & qwen2.5-3b & 0.785 & 0.823 & 0.785 & \textbf{0.868} & \textbf{0.883} & \textbf{0.868} \\
Penguins in a Table & qwen2.5-7b & 0.896 & 0.898 & 0.896 & \textbf{0.951} & \textbf{0.955} & \textbf{0.951} \\
TableBench & qwen2.5-3b & \textbf{0.323} & 0.467 & \textbf{0.346} & 0.291 & \textbf{0.471} & 0.317 \\
TableBench & qwen2.5-7b & 0.323 & 0.459 & 0.349 & \textbf{0.354} & \textbf{0.557} & \textbf{0.393} \\
\bottomrule
\end{tabular}
\caption{Additional comparison with AutoPrep, a question-aware data-preparation framework for TQA, on representative Qwen2.5 backbones.}
\label{tab:appendix_autoprep}
\end{table*}

\subsection{Additional Baseline Comparisons}
\label{appendix:other_baselines}

\label{app:additional_baselines}

We provide additional comparisons with related TableQA and multi-agent baselines that were not included in the main result tables due to differences in implementation assumptions and supported backbones. We consider two groups of baselines. First, we compare with ReAcTable and Chameleon, whose official implementations are designed around GPT-series APIs. Second, we compare with AutoPrep, a multi-agent framework for question-aware data preparation in TableQA.

\paragraph{GPT-based tool-use baselines.}
ReAcTable \cite{ReAcTable} extends the ReAct paradigm for TableQA by integrating LLM reasoning with external SQL and Python execution. It decomposes complex questions into iterative reasoning steps, executes code to obtain intermediate results, and uses these results to guide subsequent reasoning. Chameleon \cite{chameleon} is a plug-and-play compositional reasoning framework that uses an LLM-based planner to orchestrate external tools such as Python programs and other modules.

The official implementations of both baselines are designed to work with GPT-series models. Accordingly, we evaluate ReAcTable, Chameleon, and MATA using GPT-4o as the shared backbone. As shown in Table~\ref{tab:gpt4o}, MATA substantially outperforms both baselines on TableBench, where tables and questions are more complex. On Penguins in a Table, Chameleon remains competitive and obtains a slightly higher F1 score, while MATA achieves the best EM and fuzzy scores.

\paragraph{AutoPrep comparison.}
AutoPrep \cite{AutoPrep} studies a complementary problem to MATA: question-aware data preparation for TQA. Given a table and a natural language question, AutoPrep first plans high-level data-preparation operations, then generates low-level executable code for these operations, and finally executes and debugs the code to produce a prepared table before answering. Thus, unlike MATA, which focuses on adaptive reasoning-path orchestration and answer selection, AutoPrep focuses on preparing the table so that downstream TQA methods can answer more reliably.

Because AutoPrep was originally evaluated under different benchmark and backbone settings, we treat this comparison as an additional adaptation study rather than a primary baseline comparison. We evaluate AutoPrep and MATA on our benchmarks using representative Qwen2.5 backbones and the same evaluation metrics. As shown in Table~\ref{tab:appendix_autoprep}, MATA performs better in three of the four benchmark--backbone settings. The exception is TableBench with qwen2.5-3B, where AutoPrep achieves higher EM and F1, while MATA obtains slightly higher fuzzy matching. Overall, these additional results suggest that MATA's gains are not limited to the three main baselines, while also showing that question-aware data preparation can be a competitive alternative in some low-resource backbone settings.

\begin{figure*}[h!]
\centering
\includegraphics[width=\linewidth]{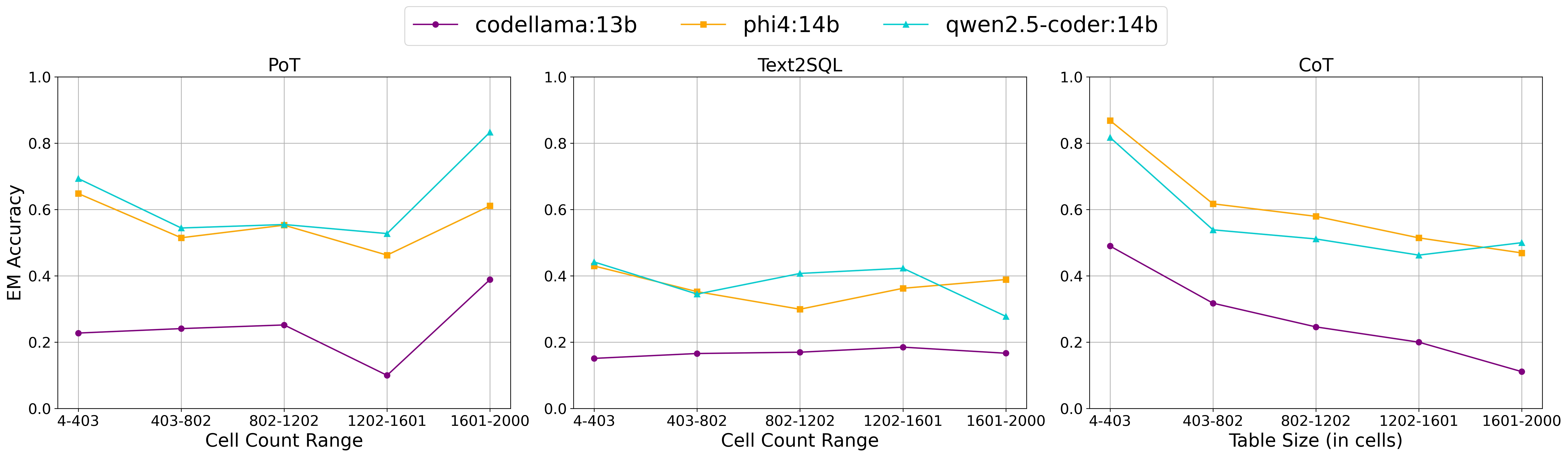} 
\caption{EM accuracy trends across varying table sizes. The horizontal axis indicates cell count ranges, and the vertical axis shows the average EM accuracy achieved by each method, illustrating how table size affects results.}
\label{fig:tablesize}
\end{figure*}

\subsection{Analysis about Table Size}
\label{appendix:sectionF}

We also aim to examine how each method is affected by the size of the input table. As shown in Figure~\ref{fig:tablesize}, CoT's EM accuracy gradually declines as the table size increases, whereas PoT and text2SQL are less sensitive to the table size. These results confirm that the table size significantly influences accuracy across methods, as shown in previous studies \cite{Mix-SC, zhang-etal-2024-syntqa}.

\newpage

\vspace*{\fill}
\null

\newpage

\section{Prompts for Experiments and Training Data Generation}
\label{appendix:sectionG}

\subsection{CoT Prompt}

\begin{tcolorbox}[breakable, colback=gray!10, colframe=black, title=CoT Agent (\textit{CoTA}) Prompt]
  Read the following table and then write solution texts to answer a question:
  \\| Name     | Number of coins | 
  \\|----------|-----------------| 
  \\| Braden   | 76              | 
  \\| Camilla  | 94              | 
  \\| Rick     | 86              | 
  \\| Mary     | 84              | 
  \\| Hector   | 80              | 
  \\| Devin    | 83              | 
  \\| Emily    | 82              | 
  \\| Avery    | 87              | 

  Question: Some friends discussed the sizes of their coin collections. What is the mean of the numbers? \\
  \#\# Return a query for the solution and answer with two keys: solution and answer. Respond using JSON only. \\

  \{\{'solution': "To find the mean, we sum up the numbers of coins for all the individuals and then divide the total by the number of individuals.
  Step 1: Add all the numbers of coins: 76 + 94 + 86 + 84 + 80 + 83 + 82 + 87 = 672
  Step 2: Count the number of individuals: There are 8 individuals.
  Step 3: Calculate the mean: Mean = Total sum of coins / Number of individuals = 672/8 = 84", 'answer': 84\}\}

  \vspace{\baselineskip} 
  
  Read the following table and then write solution texts to answer a question:
  \\| Price  | Quantity demanded | Quantity supplied | 
  \\|--------|-------------------|-------------------|
  \\| \$155   | 22,600            | 5,800            |
  \\| \$275   | 20,500            | 9,400            |
  \\| \$395   | 18,400            | 13,000           |
  \\| \$515   | 16,300            | 16,600           |
  \\| \$635   | 14,200            | 20,200           |
 
  Question: Look at the table. Then answer the question. At a price of \$155, is there a shortage or a surplus?\\
  \#\# Return a query for the solution and answer with two keys: solution and answer. Respond using JSON only.\\

  \{\{'solution': "To determine if there is a 'shortage' or a 'surplus', we compare the 'quantity demanded' and the 'quantity supplied' at the given price.
  Step 1: 
  Identify the values from the table for \$155:
  - Quantity demanded = 22,600
  - Quantity supplied = 5,800
  Step 2: 
  Calculate the difference: 
  Shortage or Surplus = Quantity demanded - Quantity supplied = 22,600 - 5,800 = 16,800
  Step 3: Determine the situation: 
  Since the quantity demanded is greater than the quantity supplied, there is a 'shortage' of 16,800 units.", 'answer': 'shortage'\}\}
  
  \vspace{\baselineskip} 
  
  Read the following table and then write solution texts to answer a question:
  \\| Name     | Number |
  \\|----------|--------|
  \\| Samir    | 7      |
  \\| Kristen  | 4      |
  \\| Dakota   | 7      |
  \\| Jamie    | 8      |
  \\| Maggie   | 9      |

  Question: Samir's class recorded how many cans of food each student collected for their canned food drive. What is the median of the numbers? \\
  \#\# Return a query for the solution and answer with two keys: solution and answer. Respond using JSON only. \\

    \{\{'solution': "To find the median, we need to arrange the numbers in order from smallest to largest and then identify the middle number.
  Step 1: Arrange the numbers: 4, 7, 7, 8, 9
  Step 2: Identify the middle number. Since there are 5 numbers (an odd count), the middle number is the third number. So the median is 7.", 'answer': 7\}\}

  \vspace{\baselineskip} 
  
  Read the following table and then write solution texts to answer a question:
  \\| Toy                     | Price  |
  \\|---------|--------|
  \\| toy boat                | \$5.54  |
  \\| toy guitar              | \$8.23  |
  \\| set of juggling balls   | \$5.01  |
  \\| trivia game             | \$8.18  |
  \\| jigsaw puzzle           | \$5.30  |
  \\| toy dinosaur            | \$3.00  |

  Question: Lorenzo has \$13.50. Does he have enough to buy a toy guitar and a set of juggling balls? \\
  \#\# Return a query for the solution and answer with two keys: solution and answer. Respond using JSON only. \\
  
  \{\{'solution': "To determine if Lorenzo can afford both the toy guitar and the set of juggling balls, we need to calculate their combined cost and compare it to Lorenzo's available money.
  Step 1: Identify the prices:
  - Toy guitar = \$8.23  
  - Set of juggling balls = \$5.01  
  Step 2: Calculate the total cost:
  Total cost = 8.23 + 5.01 = 13.24
  Step 3: Compare the total cost to Lorenzo's money. Lorenzo has \$13.50, and the total cost is \$13.24. Since \$13.24 is less than \$13.50, Lorenzo does have enough money.", 'answer': 'Yes'\}\}

  \vspace{\baselineskip} 

  \{\textit{\textbf{query}}\}

\end{tcolorbox}

\subsection{PoT Prompt}
\begin{tcolorbox}[
    breakable, colback=gray!10, colframe=black, title=PoT Agent (\textit{PoTA}) Prompt]
Read the following table and then write Python code with pandas to answer a question:
\\
import pandas as pd \\
data = \{\\
'Name': ['Braden', 'Camilla', 'Rick', 'Mary', 'Hector', 'Devin', 'Emily', 'Avery'],\\
'Number of coins': [76, 94, 86, 84, 80, 83, 82, 87]\\
\} \\
df = pd.DataFrame(data)
\\
Question: Some friends discussed the sizes of their coin collections. What is the mean of the numbers? \\
\#\# You don’t need to reprint pre-written code like 'import pandas as pd', 'data = \{...\}', or 'df = pd.DataFrame(data)'. That code will be provided separately, so just give me the code that processes 'data' and 'df'. \\
\#\# Return a query for the 'python code with pandas which return ans' with one key: code. Respond using JSON only. \\

\{\{'code' : '''\# Calculate the mean of the 'Number of coins' column\\
mean\_coins = df['Number of coins'].mean()\\
ans = mean\_coins'''\}\}

\vspace{1em}

Read the following table and then write Python code with pandas to answer a question:

import pandas as pd \\
data = \{\\
\ \ \ \ 'Price': [155, 275, 395, 515, 635],\\
\ \ \ \ 'Quantity demanded': [22600, 20500, 18400, 16300, 14200],\\
\ \ \ \ 'Quantity supplied': [5800, 9400, 13000, 16600, 20200]\\
\} \\
df = pd.DataFrame(data)

Question: Look at the table. Then answer the question. At a price of \$155, is there a shortage or a surplus? \\
\#\# You don’t need to reprint pre-written code like 'import pandas as pd', 'data = \{...\}', or 'df = pd.DataFrame(data)'. That code will be provided separately, so just give me the code that processes 'data' and 'df'. \\
\#\# Return a query for the 'python code with pandas which return ans' with one key: code. Respond using JSON only. \\

\{\{'code' : '''\# Filter the row where the price is \$155\\
price\_155 = df[df['Price'] == 155]\\
\# Calculate shortage or surplus\\
quantity\_demanded = price\_155['Quantity demanded'].values[0]\\
quantity\_supplied = price\_155['Quantity supplied'].values[0]\\
if quantity\_demanded > quantity\_supplied:\\
\ \ \ \ ans = 'shortage'\\
else:\\
\ \ \ \ ans = 'surplus' '''\}\}

\vspace{1em}

Read the following table and then write Python code with pandas to answer a question:

import pandas as pd \\
data = \{\\
\ \ \ \ 'Name': ['Samir', 'Kristen', 'Dakota', 'Jamie', 'Maggie'],\\
\ \ \ \ 'Cans collected': [7, 4, 7, 8, 9]\\
\} \\
df = pd.DataFrame(data)

Question: Samir's class recorded how many cans of food each student collected for their canned food drive. What is the median of the numbers? \\
\#\# You don’t need to reprint pre-written code like 'import pandas as pd', 'data = \{...\}', or 'df = pd.DataFrame(data)'. That code will be provided separately, so just give me the code that processes 'data' and 'df'. \\
\#\# Return a query for the 'python code with pandas which return ans' with one key: code. Respond using JSON only. \\

\{\{'code' : '''\# Calculate the median of the 'Cans collected' column\\
median\_cans = df['Cans collected'].median()\\
ans = median\_cans'''\}\} \\

\vspace{1em}

Read the following table and then write Python code with pandas to answer a question:

import pandas as pd \\
data = \{\\
\ \ \ \ 'Toy': ['toy boat', 'toy guitar', 'set of juggling balls', 'trivia game', 'jigsaw puzzle', 'toy dinosaur'],\\
\ \ \ \ 'Price': [5.54, 8.23, 5.01, 8.18, 5.30, 3.00]\\
\} \\
df = pd.DataFrame(data)

Question: Lorenzo has \$13.50. Does he have enough to buy a toy guitar and a set of juggling balls? \\
\#\# You don’t need to reprint pre-written code like 'import pandas as pd', 'data = \{...\}', or 'df = pd.DataFrame(data)'. That code will be provided separately, so just give me the code that processes 'data' and 'df'. \\
\#\# Return a query for the 'python code with pandas which return ans' with one key: code. Respond using JSON only. \\

\{\{'code' : '''\# Lorenzo's total money\\
total\_money = 13.50 \\

\# Filter the prices of 'toy guitar' and 'set of juggling balls'\\
selected\_items = df[df['Toy'].isin(['toy guitar', 'set of juggling balls'])] \\

\# Calculate the total cost\\
total\_cost = selected\_items['Price'].sum() \\

\# Determine if Lorenzo has enough money\\
if total\_money >= total\_cost:\\
\ \ \ \ ans = "yes"\\
else:\\
\ \ \ \ ans = "no"\\
'''\}\} 

\vspace{1em}

  \{\textbf{\textit{query}}\}
\end{tcolorbox}

\begin{tcolorbox}[breakable, colback=gray!10, colframe=black, title=Python Debug Agent (\textit{PDA}) Prompt]
You are an expert in reviewing and correcting Python code designed to solve questions about tables. \\
Review the query, the previous pandas code written to address it, and its execution results to identify any parts that need correction. \\

\#\#\# query \\
Read the following table and then write Python code with pandas to answer a question:

import pandas as pd \\
data = \{\\
\ \ \ \ 'Toy': ['toy boat', 'toy guitar', 'set of juggling balls', 'trivia game', 'jigsaw puzzle', 'toy dinosaur'],\\
\ \ \ \ 'Price': [5.54, 8.23, 5.01, 8.18, 5.30, 3.00]\\
\} \\
df = pd.DataFrame(data)

Question: What is the average price of toys that cost more than \$5? \\
\#\# You don’t need to reprint pre-written code like `import pandas as pd', `data = \{...\}', or `df = pd.DataFrame(data)'. That code will be provided separately, so just give me the code that processes `data' and `df'. \\
\#\# Return a query for the python code with pandas which return ans with one key: code. Respond using JSON only. (You must return the value with `ans’) \\

\#\#\# Previous Code: \\
\# The following 4 toys are included : \\
df\_previous = df[df[`Toy'].isin([\\
\ \ \ \ `toy boat', \\
\ \ \ \ `toy guitar', \\
\ \ \ \ `set of juggling balls', \\
\ \ \ \ `toy dinosaur'\\
])] \\
\# Summing the prices \\
total\_previous = df\_previous[`Price'].sum() \\
\# Counting the number of toys \\
count\_previous = len(df\_previous) \\
\# Calculating the average \\
ans = total\_previous / count\_previous \\

\#\#\# Previous Execution Result: \\
5.445 \\

\#\#\# Return a query for `corrected python code with pandas which return ans' with one key: code. Respond using JSON only. (You must return the value with 'ans') \\
\{\{`code' : ```\\
\# The toys that will actually be included among those priced greater than \$5 are the toy boat, toy guitar, set of juggling balls, trivia game, and jigsaw puzzle.\\
df\_corrected = df[df[`Toy'].isin([\\
\ \ \ \ `toy boat',\\
\ \ \ \ `toy guitar',\\
\ \ \ \ `set of juggling balls',\\
\ \ \ \ `trivia game',\\
\ \ \ \ `jigsaw puzzle'\\
])]\\
\# Summing the prices\\
total\_corrected = df\_corrected[`Price'].sum()\\
\# Counting the number of toys\\
count\_corrected = len(df\_corrected)\\
\# Calculating the average\\
ans = total\_corrected / count\_corrected\\
'''\}\} \\

\#\#\# query \\
\{\textbf{\textit{query}}\} \\

\#\#\# Previous Code: \\
\{\textbf{\textit{code}}\} \\

\#\#\# Previous Execution Result: \\
\{\textbf{\textit{execution\_result}}\} \\

\#\#\# Return a query for `corrected python code with pandas which return ans' with one key: code. Respond using JSON only. (You must return the value with `ans')

\end{tcolorbox}

\subsection{Text2SQL Prompt}
\label{subsec:text2sqlprompt}

\begin{tcolorbox}[breakable, colback=gray!10, colframe=black, title=text2SQL Agent (\textit{t2SA}) Prompt]
Read the following table and then write SQL code to answer the question:

-- Table: coin\_collection \\
-- Columns: \\
-- \ \ Name (TEXT) \\
-- \ \ Number\_of\_coins (INTEGER) \\
-- \\
-- Rows: \\
-- \ \ Braden \ \ \ \ \ \ | 76 \\
-- \ \ Camilla \ \ \ \ \ | 94 \\
-- \ \ Rick \ \ \ \ \ \ \ \ | 86 \\
-- \ \ Mary \ \ \ \ \ \ \ \ | 84 \\
-- \ \ Hector \ \ \ \ \ | 80 \\
-- \ \ Devin \ \ \ \ \ \ | 83 \\
-- \ \ Emily \ \ \ \ \ \ | 82 \\
-- \ \ Avery \ \ \ \ \ \ | 87

Question: Some friends discussed the sizes of their coin collections. What is the mean of the numbers? \\
\#\# Return a query for the `SQL code' with one key: code. Respond using JSON only.

\{\{`code' : ```\\
SELECT AVG(Number\_of\_coins) AS answer\\
FROM coin\_collection;\\
'''\}\}

\vspace{1em}

Read the following table and then write SQL code to answer the question:

-- Table: market \\
-- Columns: \\
-- \ \ Price (INTEGER) \\
-- \ \ Quantity\_demanded (INTEGER) \\
-- \ \ Quantity\_supplied (INTEGER) \\
-- \\
-- Rows: \\
-- \ \ 155 \ \ \ | 22600 | 5800 \\
-- \ \ 275 \ \ \ | 20500 | 9400 \\
-- \ \ 395 \ \ \ | 18400 | 13000 \\
-- \ \ 515 \ \ \ | 16300 | 16600 \\
-- \ \ 635 \ \ \ | 14200 | 20200

Question: Look at the table. Then answer the question. At a price of \$155, is there a shortage or a surplus? \\
\#\# Return a query for the `SQL code' with one key: code. Respond using JSON only.

\{\{`code' : ```\\
SELECT \\ 
\ \ CASE \\ 
\ \ \ \ WHEN Quantity\_demanded > Quantity\_supplied THEN `shortage'\\
\ \ \ \ ELSE `surplus'\\
\ \ END AS answer\\
FROM market\\
WHERE Price = 155;\\
'''\}\}

\vspace{1em}

Read the following table and then write SQL code to answer the question:

-- Table: can\_collection \\
-- Columns: \\
-- \ \ Name (TEXT) \\
-- \ \ Cans\_collected (INTEGER) \\
-- \\
-- Rows: \\
-- \ \ Samir \ \ \ \ \ | 7 \\
-- \ \ Kristen \ \ \ | 4 \\
-- \ \ Dakota \ \ \ \ | 7 \\
-- \ \ Jamie \ \ \ \ \ | 8 \\
-- \ \ Maggie \ \ \ | 9

Question: Samir's class recorded how many cans of food each student collected for their canned food drive. What is the median of the numbers? \\
\#\# Return a query for the `SQL code' with one key: code. Respond using JSON only.

\{\{`code' : ```\\
SELECT PERCENTILE\_CONT(0.5) WITHIN GROUP (ORDER BY Cans\_collected) AS answer\\
FROM can\_collection;\\
'''\}\}

\vspace{1em}

Read the following table and then write SQL code to answer the question:

-- Table: toys \\
-- Columns: \\
-- \ \ Toy (TEXT) \\
-- \ \ Price (DECIMAL) \\
-- \\
-- Rows: \\
-- \ \ toy boat \ \ \ \ \ \ \ \ \ \ \ \ \ \ | 5.54 \\
-- \ \ toy guitar \ \ \ \ \ \ \ \ \ \ \ \ | 8.23 \\
-- \ \ set of juggling balls | 5.01 \\
-- \ \ trivia game \ \ \ \ \ \ \ \ \ | 8.18 \\
-- \ \ jigsaw puzzle \ \ \ \ \ \ | 5.30 \\
-- \ \ toy dinosaur \ \ \ \ \ \ | 3.00

Question: Lorenzo has \$13.50. Does he have enough to buy a toy guitar and a set of juggling balls? \\
\#\# Return a query for the 'SQL code' with one key: code. Respond using JSON only.

\{\{`code' : ```\\
SELECT \\ 
\ \ CASE \\ 
\ \ \ \ WHEN SUM(Price) <= 13.50 THEN `yes'\\
\ \ \ \ ELSE `no'\\
\ \ END AS answer\\
FROM toys\\
WHERE Toy IN (`toy guitar', `set of juggling balls');\\
'''\}\}

\vspace{1em}

\{\textbf{\textit{query}}\}

\end{tcolorbox}

\begin{tcolorbox}[breakable, colback=gray!10, colframe=black, title=SQL Debug Agent (\textit{SDA}) Prompt]
You are an expert in reviewing and correcting SQL code designed to solve questions about tables. \\
Review the query, the previous SQL code written to address it, and its execution results to identify any parts that need correction. \\

\#\#\# query \\
Read the following table and then write SQL code to answer a question:

-- Table: toys \\
-- Columns: \\
-- \ \ Toy (TEXT) \\
-- \ \ Price (DECIMAL) \\

-- Rows: \\
-- \ \ toy boat \ \ \ \ \ \ \ \ \ \ \ \ \ \ \ \ \ \ \ | 5.54 \\
-- \ \ toy guitar \ \ \ \ \ \ \ \ \ \ \ \ \ \ \ \ \ | 8.23 \\
-- \ \ set of juggling balls | 5.01 \\
-- \ \ trivia game \ \ \ \ \ \ \ \ \ \ \ \ \ \ \ \ \ \ \ | 8.18 \\
-- \ \ jigsaw puzzle \ \ \ \ \ \ \ \ \ \ \ \ \ \ \ \ \ | 5.30 \\
-- \ \ toy dinosaur \ \ \ \ \ \ \ \ \ \ \ \ \ \ \ \ \ | 3.00

Question: What is the average price of toys that cost more than \$5? \\
\#\# Return a query for the 'SQL code' with one key: code. Respond using JSON only.

\vspace{1em}

\#\#\# Previous Code: \\
-- The following 4 toys are included: \\
SELECT AVG(Price) AS ans \\
FROM toys \\
WHERE Toy IN (`toy boat', `toy guitar', `set of juggling balls', `toy dinosaur');

\vspace{1em}

\#\#\# Previous Execution Result: \\
5.445

\vspace{1em}

\#\#\# Return a query for `corrected SQL code' with one key: code. Respond using JSON only. \\
\{\{`code' : ```\\
-- The toys that will actually be included among those priced greater than \$5 are:\\
-- toy boat, toy guitar, set of juggling balls, trivia game, jigsaw puzzle.\\
SELECT AVG(Price) AS ans\\
FROM toys\\
WHERE Price > 5;\\
'''\}\}

\vspace{1em}

\#\#\# query \\
\{\textbf{\textit{query}}\}

\vspace{1em}

\#\#\# Previous Code: \\
\{\textbf{\textit{code}}\}

\vspace{1em}

\#\#\# Previous Execution Result: \\
\{\textbf{\textit{execution\_result}}\}

\vspace{1em}

\#\#\# Return a query for `corrected SQL code' with one key: code. Respond using JSON only.

\end{tcolorbox}

\begin{tcolorbox}[breakable, colback=gray!10, colframe=black, title=Format Matcher (\textit{FM}) Prompt]
You are a helpful AI that extracts the key entities from a specific sentence. \\
For a given question, someone has provided an answer. \\
However, the answer is too long and verbose, so you need to condense it into a few short entities. Summarize the answer into a few words as entities. \\

\#\# Return a query for the solution and answer with two keys: Justification and Answer. Respond using JSON only. \\
- Question: Which country had the highest British exports in 1950, and how does it compare to its British exports in 1942? \\
- Answer: Sweden had the highest British exports in 1950 with 165.5 million Pounds, which was 72.3 million Pounds higher than its 1942 value of 93.2 million Pounds. \\
\#\# Return the final output strictly in the following JSON format. \\
\{\{`Extracted\_Answer' : `Sweden, 72.3'\}\} \\

\#\# Return a query for the solution and answer with two keys: Justification and Answer. Respond using JSON only. \\
- Question : \{\textbf{\textit{Question}}\} \\
- Answer : \{\textbf{\textit{Answer}}\} \\
\#\# Return the final output strictly in the following JSON format.

\end{tcolorbox}

\subsection{Judge Agent Prompt}
\label{app:ja_prompt}

The Judge Agent (\textit{JA}) is implemented as a tool-calling LLM agent. It receives the table, the question, and the JSON-formatted reasoning outputs from the CoT, PoT, and text2SQL branches. \textit{JA} can also call the confidence-score tool, which returns the \textit{CC} scores for the three reasoning paths. The prompt used for \textit{JA} is shown below.

% \begin{tcblisting}{
%   breakable,
%   colback=gray!10,
%   colframe=black,
%   title=Judge Agent (\textit{JA}) Prompt,
%   listing only,
%   listing options={
%     basicstyle=\small\ttfamily,
%     breaklines=true,
%     columns=fullflexible,
%     keepspaces=true
%   }
% }
\begin{tcolorbox}[breakable, colback=gray!10, colframe=black, title=Judge Agent (\textit{JA}) Prompt]
System: \\
You are a helpful AI that provides appropriate answers to users' questions. \\

Your task is to output both the Answer to a Question about a given Table and the Justification for that Answer. \\

The provided Table is as follows: \\
\{\textbf{\textit{table}}\}

The Question about the Table is as follows:
\{\textbf{\textit{question}}\}

The user will also provide 'the results of three different reasoning approaches performed by another AI to answer the Question about the Table.' You should use these as references to produce the optimal Answer.

These reasoning results will be provided in JSON format. The three reasoning approaches are detailed below:

1. reasoning using pandas (PoT): This refers to reasoning about the Question using Python's pandas library, including the generated code and its execution results. Here, N indicates the number of times the code was refined. A larger N means multiple refinements were attempted. For example, code with N=1 was obtained by refining the code with N=0.

2. reasoning using SQL (textSQL): This refers to reasoning about the Question using SQL, including the generated code and its execution results. As with PoT, N indicates the number of refinement iterations.

3. Text-based reasoning (CoT): This refers to reasoning about the Question purely in natural language and its resulting answer. CoT does not involve any refinements; therefore, N is always 0.

You must use the provided reasoning results to determine the best possible final answer. Make sure to use the `check\_confidence\_scores' tool for obtaining confidence scores for each approach. If these results alone are insufficient or ambiguous, you may optionally use the `check\_confidence\_scores' tool to obtain confidence scores for each reasoning result. However, do not rely on these confidence scores as 100\% accurate--they are only meant for reference. You may choose whether or not to use this tool.

Return the final output strictly in the following JSON format:
\{\{'Justification': 'To calculate the rate of change, subtract the number of boxes sold on Wednesday (27) from Thursday (23): 23 - 27 = -4. Since the change occurred over 1 day, the rate of change is -4 boxes per day.', 'Answer': '-4'\}\}

Human:
\{\textbf{\textit{results\_all}}\} \\

Agent scratchpad:
\{\textbf{\textit{agent\_scratchpad}}\}

\end{tcolorbox}

%%%%%%%%%%%%%%%%% 테이블들 %%%%%%%%%%%%%%%%%%%%

\begin{table*}[h]
\centering
\scalebox{1.00}{
\begin{tabular}{c|l|c|c|c||c|c|c|c}
\toprule
- &\textbf{Model} & \textbf{N = 1} & \textbf{N = 2} & \textbf{N = 3}  & \textbf{N = 4} & \textbf{N = 5} & \textbf{N = 6} & \textbf{N = 7} \\
\midrule
&
llama3.2-3b &
54 &
63 &
15  & 
5 & 
2  & 
2 & 
3 \\

&
mistral-7b & 
24 &
16 &
28  & 
22 & 
19  & 
13 & 
22 \\

small LLM &
phi4-mini-3.8b & 
107 &
14 &
7  & 
3 & 
0  & 
0 & 
13 \\

&
qwen2.5-3b & 
78 &
35 &
14  & 
6 & 
2  & 
2 & 
7 \\

&
qwen2.5-7b & 
79 &
28 &
11  & 
2 & 
0  & 
1 & 
23 \\

\midrule

&
mistral-small-24b & 
0 &
122 &
9  & 
2 & 
2  & 
0 & 
9 \\

&
cogito-32b & 
116 &
9 &
5  & 
0 & 
0  & 
0 & 
14 \\

large LLM &
qwen2.5-32b & 
105 &
31 &
4  & 
2 & 
0  & 
0 & 
2 \\

&
GPT-4o & 
82 &
46 &
11  & 
3 & 
2  & 
0 & 
0 \\

&
Claude-3.7-Sonnet & 
127 &
13 &
4  & 
0 & 
0  & 
0 & 
0 \\

\bottomrule
\end{tabular}
}
\caption{Early stopping iteration distribution for PoT debugging on the \textbf{Penguins in a Table} dataset with a maximum of $N = 7$ iterations. Most models converge within 1–2 iterations, demonstrating limited benefit from further refinement.}
\label{tab:PoTDebugStopPenguin}
\end{table*}
\begin{table*}[h]
\centering
\scalebox{1.00}{
\begin{tabular}{c|l|c|c|c||c|c|c|c}
\toprule
- &\textbf{Model} & \textbf{N = 1} & \textbf{N = 2} & \textbf{N = 3}  & \textbf{N = 4} & \textbf{N = 5} & \textbf{N = 6} & \textbf{N = 7} \\
\midrule
&
llama3.2-3b &
60 &
247 &
160  & 
77 & 
37  & 
6 & 
106 \\

&
mistral-7b & 
54 &
69 &
54  & 
30 & 
19  & 
12 & 
455 \\

small LLM &
phi4-mini-3.8b & 
203 &
237 &
78 & 
30 & 
12  & 
5 & 
128 \\

&
qwen2.5-3b & 
248 &
192 &
95 & 
39 & 
15  & 
10 & 
94 \\

&
qwen2.5-7b & 
380 &
132 &
58 & 
24 & 
7  & 
5 & 
87 \\

\midrule

&
mistral-small-24b & 
12 &
492 &
122 & 
41 & 
8  & 
3 & 
15 \\

&
cogito-32b & 
449 &
190 &
30 & 
7 & 
3  & 
0 & 
14 \\

large LLM &
qwen2.5-32b & 
356 &
223 &
80 & 
8 & 
7  & 
2 & 
17 \\

&
GPT-4o & 
504 &
131 &
15 & 
13 & 
10  & 
8 & 
12 \\

&
Claude-3.7-Sonnet & 
530 &
102 &
16 & 
14 & 
11  & 
8 & 
12 \\

\bottomrule
\end{tabular}
}
\caption{Early stopping iteration distribution for PoT debugging on the \textbf{TableBench} dataset. Compared to Table 16, more iterations are sometimes required due to the increased complexity of questions and tables.}
\label{tab:PoTDebugStopTablebench}
\end{table*}

% \newpage

\begin{table*}[h]
\centering
\scalebox{0.90}{
\begin{tabular}{c|l|c|c|c|c||c|c|c|c}
\toprule
- &\textbf{Model} & \textbf{N = 0} & \textbf{N = 1} & \textbf{N = 2} & \textbf{N = 3}  & \textbf{N = 4} & \textbf{N = 5} & \textbf{N = 6} & \textbf{N = 7} \\
\midrule
&
llama3.2-3b &
133 &
5 &
1 &
0 & 
0 & 
0 & 
0 & 
5 \\

&
mistral-7b & 
131 &
8 &
0 &
5 & 
0 & 
0 & 
0 & 
0 \\

small LLM &
phi4-mini-3.8b & 
128 &
8 &
4 &
0 & 
0 & 
0 & 
0 & 
4 \\

&
qwen2.5-3b & 
131 &
8 &
1 &
0 & 
1 & 
0 & 
0 & 
3 \\

&
qwen2.5-7b & 
140 &
3 &
0 &
1 & 
0 & 
0 & 
0 & 
0 \\

\midrule

&
mistral-small-24b & 
141 &
2 &
1 &
0 & 
0 & 
0 & 
0 & 
0 \\

&
cogito-32b & 
142 &
2 &
0 &
0 & 
0 & 
0 & 
0 & 
0 \\

large LLM &
qwen2.5-32b & 
143 &
1 &
0 &
0 & 
0 & 
0 & 
0 & 
0 \\

&
GPT-4o & 
144 &
0 &
0 &
0 & 
0 & 
0 & 
0 & 
0 \\

&
Claude-3.7-Sonnet & 
144 &
0 &
0 &
0 & 
0 & 
0 & 
0 & 
0 \\

\bottomrule
\end{tabular}
}
\caption{Early stopping iteration distribution for text2SQL debugging on the \textbf{Penguins in a Table} dataset. Most models terminate at iteration 0 or 1, indicating that debugging typically halts early for simple tasks. Unlike PoT, text2SQL begins counting from $N=0$ because, if no execution error occurs at the initial step, the loop stops without further debugging.}
\label{tab:text2SQLDebugStopPenguin}
\end{table*}
\begin{table*}[h]
\centering
\scalebox{0.90}{
\begin{tabular}{c|l|c|c|c|c||c|c|c|c}
\toprule
- &\textbf{Model} & \textbf{N = 0} & \textbf{N = 1} & \textbf{N = 2} & \textbf{N = 3}  & \textbf{N = 4} & \textbf{N = 5} & \textbf{N = 6} & \textbf{N = 7} \\
\midrule
&
llama3.2-3b &
252 &
173 &
139 &
43 & 
23 & 
6  & 
2 & 
55 \\

&
mistral-7b & 
249 &
42 &
30 &
25  & 
25 & 
10  & 
11 & 
301 \\

small LLM &
phi4-mini-3.8b & 
259 &
71 &
71 &
62 & 
43 & 
23  & 
14 & 
150 \\

&
qwen2.5-3b & 
284 &
78 &
88 &
71 & 
39 & 
19  & 
10 & 
104 \\

&
qwen2.5-7b & 
119 &
366 &
96 &
14 & 
5 & 
3  & 
2 & 
88 \\

\midrule

&
mistral-small-24b & 
343 &
145 &
77 &
30 & 
19 & 
5  & 
3 & 
71 \\

&
cogito-32b & 
422 &
107 &
70 &
33 & 
10 & 
5  & 
1 & 
45 \\

large LLM &
qwen2.5-32b & 
370 &
119 &
56 &
40 & 
24 & 
6 & 
2 & 
76 \\

&
GPT-4o & 
460 &
110 &
61 &
16 & 
14 & 
12  & 
8 & 
12 \\

&
Claude-3.7-Sonnet & 
464 &
176 &
28 &
7 & 
6 & 
5  & 
3 & 
4 \\

\bottomrule
\end{tabular}
}
\caption{Early stopping iteration distribution for text2SQL debugging on the \textbf{TableBench} dataset. While slightly more refinement is needed than in Table 6, most models still resolve within the first few iterations. Unlike PoT, text2SQL begins counting from $N=0$ because, if no execution error occurs at the initial step, the loop stops without further debugging.}
\label{tab:text2SQLDebugStopTablebench}
\end{table*}

\begin{table*}[h]
\centering
\scalebox{1.00}{
\begin{tabular}{c|l|c|c|c|c|c|c|c}
\toprule
- &\textbf{Model} & \textbf{N = 1} & \textbf{N = 2} & \textbf{N = 3}  & \textbf{N = 4} & \textbf{N = 5} & \textbf{N = 6} & \textbf{N = 7} \\
\midrule
&
llama3.2-3b &
140 &
4 &
0  & 
0 & 
0  & 
0 & 
0 \\

&
mistral-7b & 
117 &
13 &
1  & 
3 & 
0  & 
0 & 
10 \\

small LLM &
phi4-mini-3.8b & 
141 &
3 &
0  & 
0 & 
0  & 
0 & 
0 \\

&
qwen2.5-3b & 
120 &
17 &
7  & 
0 & 
0  & 
0 & 
0 \\

&
qwen2.5-7b & 
131 &
9 &
4  & 
0 & 
0  & 
0 & 
0 \\

\midrule

&
mistral-small-24b & 
144 &
0 &
0  & 
0 & 
0  & 
0 & 
0 \\

&
cogito-32b & 
135 &
9 &
0  & 
0 & 
0  & 
0 & 
0 \\

large LLM &
qwen2.5-32b & 
142 &
2 &
0  & 
0 & 
0  & 
0 & 
0 \\

&
GPT-4o & 
138 &
6 &
0  & 
0 & 
0  & 
0 & 
0 \\

&
Claude-3.7-Sonnet & 
144 &
0 &
0  & 
0 & 
0  & 
0 & 
0 \\

\bottomrule
\end{tabular}
}
\caption{Early stopping iteration distribution for CoT self-refinement on the \textbf{Penguins in a Table} dataset.
The majority of models terminate at iteration $N = 1$, showing that CoT reasoning rarely benefits from multiple rounds of refinement.}
\label{tab:CoTDebugStop_penguin}
\end{table*}

\begin{table*}[h]
\centering
\scalebox{1.00}{
\begin{tabular}{c|l|c|c|c|c|c|c|c}
\toprule
- &\textbf{Model} & \textbf{N = 1} & \textbf{N = 2} & \textbf{N = 3}  & \textbf{N = 4} & \textbf{N = 5} & \textbf{N = 6} & \textbf{N = 7} \\
\midrule
&
llama3.2-3b &
669 &
20 &
4  & 
0 & 
0  & 
0 & 
0 \\

&
mistral-7b & 
497 &
172 &
5  & 
2 & 
0  & 
0 & 
17 \\

small LLM &
phi4-mini-3.8b & 
582 &
109 &
2  & 
0 & 
0  & 
0 & 
0 \\

&
qwen2.5-3b & 
652 &
41 &
0  & 
0 & 
0  & 
0 & 
0 \\

&
qwen2.5-7b & 
621 &
70 &
2  & 
0 & 
0  & 
0 & 
0 \\

\midrule

&
mistral-small-24b & 
558 &
130 &
5  & 
0 & 
0  & 
0 & 
0 \\

&
cogito-32b & 
685 &
8 &
0  & 
0 & 
0   & 
0 & 
0 \\

large LLM &
qwen2.5-32b & 
652 &
40 &
1  & 
0 & 
0  & 
0 & 
0 \\

&
GPT-4o & 
684 &
8 &
1  & 
0 & 
0  & 
0 & 
0 \\

&
Claude-3.7-Sonnet & 
690 &
3 &
0  & 
0 & 
0  & 
0 & 
0 \\

\bottomrule
\end{tabular}
}
\caption{Early stopping iteration distribution for CoT self-refinement on the \textbf{TableBench} dataset.
Most models also stop at iteration $N = 1$, indicating that even in more complex scenarios, CoT outputs are generally stable and do not require iterative correction.}
\label{tab:CoTDebugStop_tablebench}
\end{table*}

\begin{table*}[h]
\centering
\begin{tabular}{c|l|c|c|c||c|c|c}
\toprule
- &\textbf{Model} & \textit{w/o CC} & \textbf{MATA} & \textbf{\%↓}  & \textit{w/o sch} & \textbf{MATA} & \textbf{\%↓} \\
\midrule
&
llama3.2-3b &
144 &
18 &
87.5\%  & 
714 & 
695  & 
2.66\% \\

&
mistral-7b & 
144 &
5 &
96.5\%  & 
823 & 
786  & 
4.50\% \\

small LLM &
phi4-mini-3.8b & 
144 &
7 &
95.1\%  & 
659 & 
635  & 
3.64\% \\
&
qwen2.5-3b & 
144 &
6 &
95.8\%  & 
695 & 
641  & 
7.77\% \\

&
qwen2.5-7b & 
144 &
1 &
99.3\%  & 
616 & 
527  &  
14.44\% \\

\midrule

&
mistral-small-24b & 
144 &
10 &
93.1\%  &
746  & 
584 & 
21.72\% \\

&
cogito-32b & 
144 &
1 &
99.3\% & 
625 & 
475  & 
24.00\% \\

large LLM &
qwen2.5-32b & 
144 &
2 &
98.6\%  & 
624 & 
483  & 
22.60\% \\

&
GPT-4o & 
144 &
8 &
94.4\%  & 
651 & 
495  & 
23.96\% \\

&
Claude-3.7-Sonnet & 
144 &
2 &
98.6\%  & 
597 & 
444  & 
25.63\% \\

\midrule
&
\textit{total} & 
1440 &
60 &
\textbf{95.8\%}  & 
6750 & 
5765  & 
\textbf{14.59\%} \\

\bottomrule
\end{tabular}
\caption{LLM agent call reduction analysis on the \textbf{Penguins in a Table} dataset (144 examples per model). Left (Confidence Checker Effect): In the w/o \textit{CC} setting, the Judge Agent (\textit{JA}) is mandatorily invoked for every query to select the final answer, resulting in 1,440 total calls. MATA utilizes \textit{CC} to validate candidates and bypasses \textit{JA} when confidence is high, drastically reducing \textit{JA} calls by 95.8\%. Right (Scheduler Effect): w/o \textit{sch} executes both PoT and text2SQL agents for all queries. The Scheduler prioritizes one path and skips the other if it aligns with CoT, achieving a 14.59\% reduction.}
\label{tab:schCC_penguin}
\end{table*}

% \caption{LLM agent call reduction by the Confidence Checker (\textit{CC}) and Scheduler (\textit{Sch}) on the \textbf{Penguins in a Table} dataset (144 examples per model).Each row shows the number of LLM agent calls without \textit{CC} (\textit{w/o CC}) and with MATA (which includes \textit{CC}), along with the relative reduction (\%↓). The same applies to Sch in the right half of the table. Results are averaged over 10 models. \textit{CC} achieves a 95.8\% reduction, and \textit{Sch} achieves 14.59\%, with larger gains observed for large LLMs. }

\begin{table*}[h]
\centering
\scalebox{1.00}{
\begin{tabular}{c|l|c|c|c||c|c|c}
\toprule
- &\textbf{Model} & \textit{w/o CC} & \textbf{MATA} & \textbf{\%↓}  & \textit{w/o sch} & \textbf{MATA} & \textbf{\%↓} \\
\midrule
&
llama3.2-3b &
693 &
293 &
57.7\% & 
4600 & 
4427  & 
3.76\% \\

&
mistral-7b & 
693 &
362 &
47.8\%  & 
5345 & 
5287  & 
1.09\% \\

small LLM &
phi4-mini-3.8b & 
693 &
388 &
44.0\%  & 
4455 & 
4342  & 
2.54\% \\
&
qwen2.5-3b & 
693 &
362 &
47.8\%  & 
4452 & 
4202  & 
5.62\% \\

&
qwen2.5-7b & 
693 &
350 &
49.5\%  & 
3717 & 
3627  &  
2.42\% \\

\midrule

&
mistral-small-24b & 
693 &
210 &
69.7\%  & 
4323 & 
3863  & 
10.64\% \\

&
cogito-32b & 
693 &
188 &
72.9\%  & 
3599 & 
3147  & 
12.56\% \\

large LLM &
qwen2.5-32b & 
693 &
209 &
69.8\%  & 
3897 & 
3419  & 
12.27\% \\

&
GPT-4o & 
693 &
187 &
73.0\%  & 
3436 & 
2915  & 
15.16\% \\

&
Claude-3.7-Sonnet & 
693 &
178 &
74.3\%  & 
3300 & 
2762  & 
16.30\% \\

\midrule

&
\textit{total} & 
6930 &
2727 &
\textbf{60.6\%}  & 
41124 & 
37991  & 
\textbf{7.62\%} \\

\bottomrule
\end{tabular}
}
\caption{LLM agent call reduction on the \textbf{TableBench} dataset (693 examples per model). Left (Confidence Checker Effect): Without \textit{CC} (w/o \textit{CC}), the system forces the Judge Agent (\textit{JA}) to evaluate all 6,930 instances. MATA leverages \textit{CC} to assess answer confidence and invokes JA only for uncertain cases, thereby cutting expensive JA inference by 60.6\%. Right (Scheduler Effect): Similarly, w/o \textit{sch} runs all reasoning paths, whereas the Scheduler selectively skips the redundant path (either PoT or text2SQL) when consistency is found, reducing total agent calls by 7.62\%.}

\label{tab:schCC_tablebench}
\end{table*}

% \caption{LLM agent call reduction by the Confidence Checker (\textit{CC}) and Scheduler (\textit{Sch}) on the \textbf{TableBench} dataset (693 examples per model).
% \textit{w/o CC} and \textit{w/o sch} denote the number of LLM agent calls without applying \textit{CC} and \textit{Sch}, respectively. \textbf{MATA} shows the call count with each module applied, and \%↓ indicates the relative reduction.
% Averaged over 10 models, \textit{CC} reduces LLM agent calls by 60.6\%, and \textit{Sch} by 7.62\%.
% }

%%%%%%%%%%%%%%%%%%%%%%%%%%%%%%%%%%%%%

% \newpage
% .

% \newpage

% \section{MATA Algorithm}
% \label{appendix:D}

% This section shows the entire framework process of MATA expressed in a single capital code.

\begin{algorithm*}[t!] % [h!]보다는 [t!]가 2단 편집(algorithm*)에서 더 안전합니다.
\caption{The entire inference process of our proposed MATA} \label{algo:mata}
\begin{algorithmic}
\Require Table ($T$), Question ($Q$), Refinement Count ($N$), Threshold ($\theta$)
\Ensure Final Answer ($A_f$)

\Statex\makebox[\linewidth]{\rule{\linewidth}{0.4pt}}

\Function{\textnormal{Code\&Debug}}{$\textit{Agent}, T, Q, N$}
    \State Initialize $\mathcal{C}, A \gets \emptyset$
    \If {$\textit{Agent}$ == \textit{PoTA}}
        \State $\textit{Debug}$ $\gets$ \textit{PDA}
    \Else \State $\textit{Debug}$ $\gets$ \textit{SDA}
    \EndIf

    \State $code^0$, $A^0$ $\gets$ $\textit{Agent}(T, Q)$
    \State Append $(code^0, A^0)$ to $\mathcal{C}, A$
    \For{$i = 0$ to $N-1$}
        \State $code^{i+1}$, $A^{i+1}$ $\gets$ $\textit{Debug}(T, Q, code^i, A^i)$
        \State Append $(code^{i+1}, A^{i+1})$ to $\mathcal{C}, A$
        \If{$\mathrm{Stop\_condition}$ == True}
            \State \textbf{break}
        \EndIf
    \EndFor
    \State \Return $\mathcal{C}$, $A$
\EndFunction

\\

\State $sol_{cot}$, $A_{cot}$ $\gets$ $\textit{CoTA}(T, Q)$

\If{Use\_Scheduler == True}
    \State prob\_pot, prob\_sql $\gets$ $\textit{Sch}(T, Q)$
    \If{prob\_pot \textgreater= prob\_sql}
        \State $\mathcal{C}_{pot}, A_{pot} \gets$ $\text{Code\&Debug}$(\textit{PoTA}, $T$, $Q$, $N$)
        \If{$A_{cot} \neq$ Last Answer in $\mathcal{A}_{pot}$}
            \State $\mathcal{C}_{sql}, A_{sql} \gets$ $\text{Code\&Debug}$(\textit{t2SA}, $T$, $Q$, $N$)
        \Else
            \State $\mathcal{C}_{sql}, A_{sql} \gets$ $\emptyset$
        \EndIf
    \Else
        \State $\mathcal{C}_{sql},  A_{sql} \gets$ $\text{Code\&Debug}$(\textit{t2SA}, $T$, $Q$, $N$)
        \If{$A_{cot} \neq$ Last Answer in $A_{sql}$}
            \State $\mathcal{C}_{pot}, A_{pot} \gets$ $\text{Code\&Debug}$(\textit{PoTA}, $T$, $Q$, $N$)
        \Else
            \State $\mathcal{C}_{pot}, A_{pot} \gets$ $\emptyset$
        \EndIf
    \EndIf
\Else
    \State $\mathcal{C}_{pot}, A_{pot} \gets$ $\text{Code\&Debug}$(\textit{PoTA}, $T$, $Q$, $N$)
    \State $\mathcal{C}_{sql},  A_{sql} \gets$ $\text{Code\&Debug}$(\textit{t2SA}, $T$, $Q$, $N$)
\EndIf

\\

\State C, P, TS $\gets$ \textit{CC}($T$, $Q$, $sol_{cot}$, $A_{cot}$, $\mathcal{C}_{pot}$, $A_{pot}$, $\mathcal{C}_{sql}$, $A_{sql}$)

\If {max(C, P, TS) $> \theta$}
    \State $A_f$ $\gets \arg\max_{A \in \{A_{cot}, A_{pot}, A_{sql}\}} \{$C, P, TS$\}$
\Else 
    \State $A_f$ $\gets$ \textit{JA}($T$, $Q$, $sol_{cot}$, $A_{cot}$, $\mathcal{C}_{pot}$, $A_{pot}$, $\mathcal{C}_{sql}$, $A_{sql}$)
\EndIf

\If {len($A_f$) $> 100$}
    \State $A_f$ $\gets$ \textit{FM}($A_f$)
\EndIf

\State \Return $A_f$
\end{algorithmic}
\end{algorithm*}

\end{document}